\pdfoutput=1

\documentclass[11pt]{article}

\usepackage{ACL2023}

\usepackage{times}
\usepackage{latexsym}

\usepackage[T1]{fontenc}

\usepackage[utf8]{inputenc}

\usepackage{microtype}

\usepackage{inconsolata}
\usepackage{amsmath}
\usepackage{bm}
\usepackage{url}
\usepackage{booktabs}
\usepackage{graphicx}
\usepackage{multirow}
\usepackage{multicol}
\usepackage{amssymb}
\usepackage{makecell}
\usepackage[ruled,linesnumbered,commentsnumbered]{algorithm2e}

\title{Back Translation for Speech-to-text Translation Without Transcripts}

\author{
    Qingkai Fang$^{1,2}$,
    Yang Feng$^{1,2}$\thanks{ $\;\;$Corresponding author: Yang Feng.} \\
    \textsuperscript{\rm1}Key Laboratory of Intelligent Information Processing \\ Institute of Computing Technology, Chinese Academy of Sciences (ICT/CAS) \\
    \textsuperscript{\rm2}University of Chinese Academy of Sciences, Beijing, China \\
    \texttt{\{fangqingkai21b, fengyang\}@ict.ac.cn} \\
}

\begin{document}
\maketitle
\begin{abstract}
The success of end-to-end speech-to-text translation (ST) is often achieved by utilizing \emph{source transcripts}, \emph{e.g.}, by pre-training with automatic speech recognition (ASR) and machine translation (MT) tasks, or by introducing additional ASR and MT data. Unfortunately, transcripts are only sometimes available since numerous unwritten languages exist worldwide. In this paper, we aim to utilize large amounts of target-side monolingual data to enhance ST without transcripts. Motivated by the remarkable success of back translation in MT, we develop a back translation algorithm for ST (\textbf{\textsc{BT4ST}}) to synthesize pseudo ST data from monolingual target data. To ease the challenges posed by short-to-long generation and one-to-many mapping, we introduce self-supervised discrete units and achieve back translation by cascading a \emph{target-to-unit} model and a \emph{unit-to-speech} model. With our synthetic ST data, we achieve an average boost of 2.3 BLEU on MuST-C En$\rightarrow$De, En$\rightarrow$Fr, and En$\rightarrow$Es datasets. More experiments show that our method is especially effective in low-resource scenarios.\footnote{Code is publicly available at \url{https://github.com/ictnlp/BT4ST}.}\footnote{Examples of synthetic ST data are available at \url{https://bt4st.github.io/} and Appendix \ref{sec:case}.}

\end{abstract}

\section{Introduction}
\label{sec:intro}

End-to-end speech-to-text translation (ST) means directly translating speech in the source language to target text without generating source transcripts~\citep{berard2016listen, duong-etal-2016-attentional}. Different from traditional cascading methods which first transcribe the speech with automatic speech recognition (ASR) and then translate the transcripts into the target text with machine translation (MT), end-to-end ST has the potential to reduce latency and avoid error propagation. Hence, it has drawn much attention and achieved great success in recent years~\citep{anastasopoulos-etal-2021-findings, anastasopoulos-etal-2022-findings}.

However, it is challenging to train an end-to-end ST model with only speech-translation pairs. Traditional cascaded models learn cross-modal mapping with ASR and cross-lingual mapping with MT.
In contrast, end-to-end ST requires simultaneous cross-modal and cross-lingual mapping, which is more complicated and usually relies on more training data. However, the amount of ST data is usually limited due to the high cost of data collection, so the ST model trained with only speech-translation pairs is usually unsatisfactory.


To tackle these problems, researchers often utilize \emph{source transcripts} to assist ST training by introducing auxiliary ASR and MT tasks. With abundant ASR and MT data, the ASR task can help the model learn better cross-modal mapping, while the MT task can help learn better cross-lingual mapping, which can significantly improve ST as shown in recent ST studies~\citep{wang2020bridging, xu-etal-2021-stacked, ye2021end, fang-etal-2022-stemm}. Unfortunately, source transcripts are not always available. 
It is estimated that there are around 3000 unwritten languages in the world which have no orthography for transcription\footnote{\url{https://www.ethnologue.com/}}. For those languages, we can no longer leverage source transcripts to help with training, so many of the latest techniques fail to benefit them.


How to train a stronger ST model without transcripts? In this paper, we address this question from the perspective of data augmentation. Although ASR and MT data are unavailable, a large amount of target-side monolingual data is still easily accessible. Motivated by the success of back translation in MT~\citep{sennrich-etal-2016-improving, edunov-etal-2018-understanding}, we aim to develop a back translation algorithm for ST (\textbf{\textsc{BT4ST}}) to synthesize pseudo ST data from monolingual target data. However, compared to text-to-text back translation, generating source speech from the target text without source transcripts is much more challenging\footnote{When transcripts are available, we can decompose it into two sub-tasks: MT for generating transcripts and text-to-speech (TTS) for speech synthesis, which becomes much more manageable.}. First, the length of text is usually only tens or hundreds, but the length of speech is about tens of thousands\footnote{For 16kHz audio waveform, 1 second of speech corresponds to a sequence of 16,000 samples.}. Therefore, the model is required to generate an extremely long sequence from a short sequence \emph{without} the assumption of monotonic alignment, which is a more difficult sequential decision problem. Second, the conversion from text to speech is a one-to-many mapping problem due to the variations in speech~\citep{chen2021adaspeech, ren2021fastspeech}. For example, the pronunciation of the same content may differ among speakers.

To address these challenges, we introduce self-supervised discrete units of the source speech as intermediate representations, and achieve back translation by cascading a \emph{target-to-unit} model and a \emph{unit-to-speech} model. Since the length of unit sequences is similar to the length of target characters\footnote{1 second of speech corresponds to 50 discrete units.}, and there is a monotonic assumption between the unit sequence and the source speech, the short-to-long generation problem in \emph{target-to-unit} and \emph{unit-to-speech} models can be greatly alleviated. Besides, discrete units can disentangle content information from other variation information (\emph{e.g.}, pitch and speaker)~\citep{polyak21_interspeech}, which eases the one-to-many mapping problem in the \emph{target-to-unit} model\footnote{It should be noted that the one-to-many mapping problem still exist due to the many-to-many mappings between transcripts and translations, but it is not the focus of this work.}. We also introduce a speaker encoder to provide speaker information as input to the \emph{unit-to-speech} model, which allows us to generate diverse source speeches by coupling different speaker representations. Following this pipeline, we can synthesize large amounts of pseudo ST data from monolingual target data without requiring transcripts. Finally, we train our ST model with both synthetic and real data.

We conduct experiments on MuST-C En$\rightarrow$De, En$\rightarrow$Fr, and En$\rightarrow$Es datasets. By leveraging about 5M additional monolingual target data for each language pair, we achieve an average improvement of 2.3 BLEU compared with the strong baseline. We also observe that our approach is more effective in low-resource scenarios, yielding a boost of 5.6 BLEU when only 100 hours of ST data are available. In addition, we generate multiple different pseudo datasets with a simple  \textbf{\emph{Diverse} \textsc{BT4ST}} method, train several models separately with each dataset, and ensemble them together. By ensembling five models, we achieve an average boost of 4.0 BLEU in three translation directions.

\section{Background}
Our work focuses on developing a back translation algorithm for ST. We first introduce the task definition and model architecture of ST in Section \ref{sec:st}, and then introduce the concept of back translation in Section \ref{sec:bt}.

\subsection{Speech-to-text Translation}
\label{sec:st}

\paragraph{Task Definition}
The goal of speech-to-text translation (ST) is to translate speech in one language into text in another language. We denote the source speech as $\mathbf{x}=(x_1, ..., x_I)$, where $I$ is the length of the audio waveform. The model generates the target sentence $\mathbf{y}=(y_1, ..., y_J)$, where $J$ is the length of the target text. In this paper, we assume that transcripts of the source speech are not available, which is a more general scenario considering about 3000 unwritten languages in the world.

\paragraph{Model Architecture}
Our ST model consists of three stacked modules: \emph{acoustic encoder}, \emph{length adaptor}, and \emph{translation model}. The \emph{acoustic encoder} is a HuBERT~\citep{hubert} model pre-trained on unlabelled audio data, which can generate meaningful representations for the source speech. The \emph{length adaptor}~\citep{li-etal-2021-multilingual} is a series of convolutional layers to shrink the length of speech representations by a factor of 4. The \emph{translation model} is a Transformer~\citep{transformer} with $N$ encoder layers and $N$ decoder layers, which takes the shrunken speech representations as input and outputs the target sentence. We train the ST model by minimizing the cross-entropy loss:
\begin{equation}
    \mathcal{L}_{\rm ST}=-\sum_{j=1}^{J}\log p(y_j|\mathbf{x}, \mathbf{y}_{<j}).
\end{equation}

\begin{figure*}[t]
    \centering
    \includegraphics[width=\textwidth]{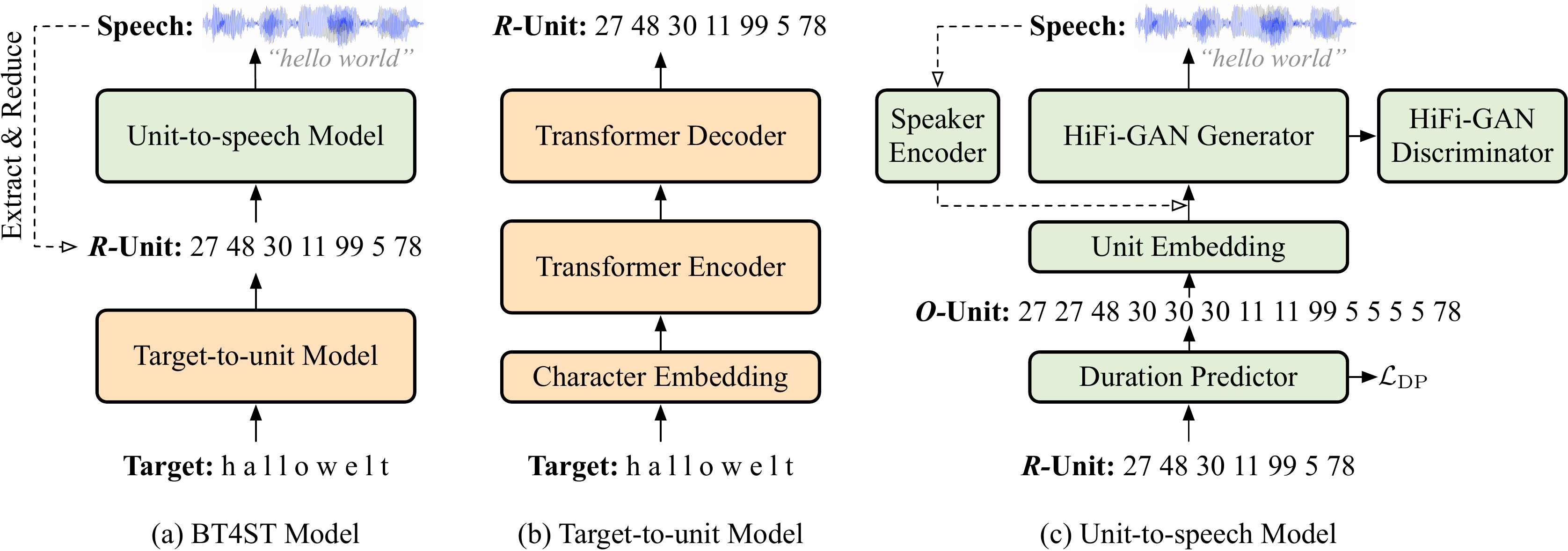}
    \caption{The overall architecture of our model. \textbf{\textit{R-}Unit}: reduced discrete units; \textbf{\textit{O-}Unit}: original discrete units.}
    \label{fig:model}
\end{figure*}

\subsection{Back Translation}
\label{sec:bt}

Back translation (BT) is a simple and effective method to leverage target-side monolingual data in neural machine translation (NMT). Formally, given a parallel corpus $\mathcal{D}=\{(\mathbf{x}^{(n)}, \mathbf{y}^{(n)})\}_{n=1}^N$, and a monolingual corpus of the target language $\mathcal{T}=\{\mathbf{y}^{(m)}\}_{m=1}^{M}$, we first train a \emph{target-to-source} model on $\mathcal{D}$. Next, we use this model to generate additional pseudo parallel data $\widetilde{\mathcal{D}}=\{(\widetilde{\mathbf{x}}^{(m)}, \mathbf{y}^{(m)})\}_{m=1}^M$ from the monolingual corpus $\mathcal{T}$. Finally, $\widetilde{\mathcal{D}}$ can be used as a complement to $\mathcal{D}$ to train a stronger \emph{source-to-target} model.


\section{Method: \textsc{BT4ST}}
How to acquire a \emph{target-to-source} model for ST given a ST parallel corpus? Directly generating source speech from the target text is a challenging problem. Inspired by recent success in self-supervised discrete representation learning for speech~\citep{baevski2020wav2vec, hubert, lakhotia-etal-2021-generative}, we first transform the source speech into a sequence of discrete units with a speech pre-trained model (Section \ref{sec:unit}), which is used as an intermediate representation in the back translation process. With discrete units, we train a \emph{target-to-unit} model (Section \ref{sec:target2unit}) and a \emph{unit-to-speech} model (Section \ref{sec:unit2speech}) on the parallel corpus, where the former predicts the sequence of discrete units corresponding to the source speech, and the latter converts discrete units into waveform. The model architecture is illustrated in Figure \ref{fig:model}.

\subsection{Unit-based Speech Representation}
\label{sec:unit}

We use the pre-trained HuBERT~\citep{hubert} model to generate discrete units corresponding to the source speech following \citet{lee-etal-2022-direct, lee-etal-2022-textless}. HuBERT generates 50Hz continuous representations for the input speech. We apply the K-means clustering algorithm to the continuous representations of the training data, and then transform the continuous representations into the corresponding cluster indices, \emph{i.e.}, discrete units. Finally, the input speech $\mathbf{x}=(x_1, ..., x_I)$ is converted into a sequence of discrete units $\mathbf{z}=(z_1, ..., z_T), z_t\in\{0,1,...,K-1\},\forall 1\leq t\leq T$, where $K$ is the number of clusters, and $T$ is the number of frames where $T=\lfloor\frac{I}{320}\rfloor$. 
There are two advantages to use discrete units as intermediate representations rather than predicting audio waveform directly. First, the sequence of discrete units is much shorter than the audio waveform, alleviating the difficulty of short-to-long generation. Second, discrete units can disentangle speech content from the pitch and speaker information~\citep{polyak21_interspeech}, which eases the one-to-many mapping problem. 

\subsection{Target-to-unit Model}
\label{sec:target2unit}

Our \emph{target-to-unit} model is a Transformer-based sequence-to-sequence model, which predicts discrete units of the source speech based on the target text. The target text $\mathbf{y}$ is tokenized as characters and fed to the encoder. For the discrete units $\mathbf{z}$, we first merge repeating units into a single one to obtain the \emph{reduced} discrete units $\mathbf{z'}=(z'_1, ..., z'_{T'})$ following \citet{lee-etal-2022-direct}. For example, $(1,1,2,2,2,3,4,4)$ will collapse to $(1,2,3,4)$. We then train the model with \emph{reduced} discrete units as the target, which can accelerate training and inference. The training objective is as follows:
\begin{equation}
    \mathcal{L}_{\rm T2U}=-\sum_{t=1}^{T'}\log p(z'_t|\mathbf{y}, \mathbf{z}'_{<t}).
\end{equation}

\subsection{Unit-to-speech Model}
\label{sec:unit2speech}

Our \emph{unit-to-speech} model is a unit-based HiFi-GAN vocoder~\citep{hifi-gan} as proposed in \citet{polyak21_interspeech}. It takes the \emph{reduced} discrete units as input and generates the waveform. The model consists of four modules: \emph{duration predictor}, \emph{speaker encoder}, \emph{generator}, and \emph{discriminator}. 

\paragraph{Duration Predictor}
As the output of the \emph{target-to-unit} model is \emph{reduced} discrete units, we add a duration predictor~\citep{ren2021fastspeech} to predict the duration of each unit. It consists of two 1D-convolutional layers with ReLU activation, each followed by layer normalization and dropout. Finally, a linear layer projects the hidden state into a scalar as duration. We denote the predicted duration vector as $\mathbf{d}=(d_1, ..., d_{T'})$, and the ground truth as $\mathbf{d}^*=(d^*_1, ..., d^*_{T'})$. The duration predictor is optimized with Mean Squared Logarithmic Error (MSLE) between ground truth and predictions as:
\begin{equation}
\label{eq:dp}
    \mathcal{L}_{\rm DP} = \frac{1}{T'}\sum_{t=1}^{T'}(\log(1+d_t)-\log(1+d_t^*))^2.
\end{equation}

Given the duration vector, we expand \emph{reduced} discrete units by repeating each unit. For example, given $\mathbf{z}'=(1,2,3,4)$ and the corresponding duration vector $\mathbf{d}=(2,3,1,2)$, the expanded sequence becomes $\mathbf{z}=(1,1,2,2,2,3,4,4)$. We use ground truth duration during training and predicted one during inference. Finally, the units are converted to continuous representations via a look-up table.

\paragraph{Speaker Encoder}
Discrete units contain little speaker information, but ST corpus usually contains speech from multiple speakers. To ease the one-to-many mapping problem in speech synthesis, we introduce a speaker encoder to extract the speaker information. The speaker encoder is a pre-trained speaker verification network~\citep{dvector}, which extracts a single 256-dimensional speaker embedding from the speech. The speaker embedding is then concatenated to the representation of each unit. During inference, we randomly select a speaker embedding from the training set for each sample. It allows us to synthesize a pseudo ST dataset containing multiple speakers.

\paragraph{Generator and Discriminator}
The generator and discriminator are the same as the original HiFi-GAN~\citep{hifi-gan}. The generator consists of several stacked blocks, each containing a transposed convolution layer followed by multiple residual blocks. It takes the concatenated features as input and outputs the waveform $\mathbf{x}$. The discriminator contains a Multi-Period Discriminator (MPD) and a Multi-Scale Discriminator (MSD), which are used to identify the periodic or consecutive patterns in the audio. The generator and discriminator are trained adversarially. More details about the HiFi-GAN can be found in Appendix \ref{appendix:hifigan}.

\subsection{Data Selection and Model Training}
\label{sec:data_select}

\begin{algorithm}[t]
\caption{Back Translation for ST}\label{algo:bt4st}
  \SetKwInOut{Input}{Input}\SetKwInOut{Output}{Output}
  \SetKwProg{proc}{Procedure}{}{End}
  \SetKwFunction{bt}{BT4ST}\SetKwFunction{train}{Training}
  \Input{ST data $\mathcal{D}=\{(\mathbf{x}^{(n)}, \mathbf{y}^{(n)})\}_{n=1}^N$, \\
        Target data $\mathcal{T}=\{\mathbf{y}^{(m)}\}_{m=1}^{M}$, \\
        Data selection ratio $\rho$}
  \Output{ST model $M_{x\rightarrow y}$}
  \BlankLine
  \proc{\bt{$\mathcal{D}$, $\mathcal{T}$, $\rho$}}{
  Get \emph{reduced} discrete units $\mathbf{z}'$ from $\mathbf{x}$ to create $\mathcal{D}'=\{(\mathbf{x}^{(n)}, \mathbf{z}'^{(n)}, \mathbf{y}^{(n)})\}_{n=1}^N$, for every $(\mathbf{x}, \mathbf{y})\in\mathcal{D}$  \\
  Train \emph{target-to-unit} model $M_{z'\leftarrow y}$ and \emph{unit-to-target} model $M_{z'\rightarrow y}$ with paired data $(\mathbf{z}', \mathbf{y})$ in $\mathcal{D}'$ \\
  Train \emph{unit-to-speech} model $M_{x\leftarrow z'}$ with paired data $(\mathbf{x}, \mathbf{z}')$ in $\mathcal{D}'$ \\
  Use $M_{z'\leftarrow y}$ and $M_{x\leftarrow z'}$ to create $\widetilde{\mathcal{D}}=\{(\widetilde{\mathbf{x}}^{(m)}, \widetilde{\mathbf{z}}'^{(m)}, \mathbf{y}^{(m)})\}_{m=1}^M$, for every $\mathbf{y}\in\mathcal{T}$ \\
  Use $M_{z'\rightarrow y}$ to translate $\widetilde{\mathbf{z}}'$ into $\widetilde{\mathbf{y}}$, and compute ${\rm BLEU}(\widetilde{\mathbf{y}}, \mathbf{y})$, for every $(\widetilde{\mathbf{x}}, \widetilde{\mathbf{z}}', \mathbf{y})\in\widetilde{\mathcal{D}}$ \\
  Select top $\rho\cdot M$ samples from $\widetilde{\mathcal{D}}$ based on BLEU scores, denoted as $\widetilde{\mathcal{D}}_S$ \\
  \train{$\mathcal{D}$, $\widetilde{\mathcal{D}}_S$}
  } 
  \BlankLine
  \proc{\train{$\mathcal{D}$, $\widetilde{\mathcal{D}}_S$}}{
  Pre-train ST model $M_{x\rightarrow y}$ on $\widetilde{\mathcal{D}}_S$ \\
  Fine-tune ST model $M_{x\rightarrow y}$ on $\mathcal{D}$
  }
\end{algorithm}

By cascading the \emph{target-to-unit} model and \emph{unit-to-speech} model, we can synthesize ST data $\widetilde{\mathcal{D}}=\{(\widetilde{\mathbf{x}}^{(m)}, \widetilde{\mathbf{z}}'^{(m)}, \mathbf{y}^{(m)})\}_{m=1}^M$ from the target-side monolingual corpus $\mathcal{T}=\{\mathbf{y}^{(m)}\}_{m=1}^{M}$, where $\widetilde{\mathbf{x}}$ is the synthetic speech and $\widetilde{\mathbf{z}}'$ is the corresponding \emph{reduced} discrete units. However, the synthetic corpus may contain some low-quality data, which may hurt model training. Therefore, we introduce \emph{data selection} and \emph{2-stage model training} to better utilize the synthetic data.

\paragraph{Data Selection}
To identify the low-quality data, we train a \emph{unit-to-target} model to translate the generated units $\widetilde{\mathbf{z}}'$ back into target text $\widetilde{\mathbf{y}}$, and compute the BLEU score~\citep{papineni-etal-2002-bleu} between $\widetilde{\mathbf{y}}$ and the ground truth $\mathbf{y}$, \emph{i.e.,} ${\rm BLEU}(\widetilde{\mathbf{y}}, \mathbf{y})$. Finally, we only keep the top $\rho\cdot M$ samples according to the BLEU score, where $\rho$ is the selection ratio. The remaining low-quality samples are discarded.

\paragraph{2-stage Model Training}
After we get the selected synthetic ST data, we train the model in a 2-stage manner following \citet{taglessbt}. We first pre-train the model on synthetic data, and then fine-tune the model on real data. This can prevent the noise-infested synthetic data from overwhelming real data during training. Algorithm \ref{algo:bt4st} describes the whole process of our proposed method.

\section{Experiments}
\subsection{Datasets}
\paragraph{MuST-C}
MuST-C~\citep{di-gangi-etal-2019-must} is a multilingual speech translation dataset, which contains about 400 hours of English (En) audio clips and corresponding translations in 8 languages: German (De), French (Fr), Spanish (Es), Italian (It), Portuguese (Pt), Dutch (Nl), Romanian (Ro), and Russian (Ru). We conduct experiments on En$\rightarrow$De, En$\rightarrow$Fr, and En$\rightarrow$Es, because these three languages have larger public monolingual corpora.

\paragraph{Monolingual Target Data}
We use the target text in WMT~\citep{buck-koehn-2016-findings} En$\rightarrow$De, En$\rightarrow$Fr, and En$\rightarrow$Es datasets as monolingual target data. Specifically, we use \emph{europarl v7}\footnote{\url{http://statmt.org/wmt13/training-parallel-europarl-v7.tgz}}, \emph{commoncrawl}\footnote{\url{http://statmt.org/wmt13/training-parallel-commoncrawl.tgz}}, \emph{news commentary v12}\footnote{\url{http://data.statmt.org/wmt17/translation-task/training-parallel-nc-v12.tgz}} subsets for all three languages. The source text is never used in the experiments. The detailed statistics of data we used are shown in Table \ref{tab:dataset}.

\subsection{Model Settings}
\paragraph{Target-to-unit Model}
The \emph{target-to-unit} model is a Transformer with 6 encoder layers and 6 decoder layers. Each layer comprises 512 hidden states, 4 attention heads, and 2048 feed-forward hidden states. The dropout is 0.3, and the label smoothing is 0.1. For target text, we first lowercase the text and segment it into characters using SentencePiece\footnote{\url{https://github.com/google/sentencepiece}}. For discrete units, we use the pre-trained quantized model\footnote{\url{https://dl.fbaipublicfiles.com/textless_nlp/gslm/hubert/km100/km.bin}}, which learns $K=100$ clusters from the 6th layer representations of pre-trained HuBERT-Base model\footnote{\url{https://dl.fbaipublicfiles.com/hubert/hubert_base_ls960.pt}}, to convert the speech into discrete units. During training, the batch size is 400, and the maximum learning rate is 5e-4. We train the model up to 100k steps with Adam optimizer~\citep{adam}. During inference, we use beam search with a beam size of 8 to generate the \emph{reduced} discrete units.

The \emph{unit-to-target} model is the same as the \emph{target-to-unit} model except for the translation direction, which is used for data selection. During inference, we use greedy search to save time. 

\begin{table}[t]
    \centering
    \resizebox{\linewidth}{!}{
    \begin{tabular}{c|cc|ccc|c}
        \toprule
         \multirow{2}{*}{\textbf{Target}} & \multicolumn{2}{c|}{\textbf{MuST-C (En$\rightarrow$)}} & \multicolumn{4}{c}{\textbf{Monolingual Data}} \\
         & hours & \#samples & Euro. & CC. & NC. & All \\
        \midrule
         \textbf{De} & 408 & 234k & 1.9M & 2.4M & 0.3M & 4.6M \\
         \textbf{Fr} & 492 & 280k & 2.0M & 3.2M & 0.3M & 5.5M \\
         \textbf{Es} & 504 & 270k & 2.0M & 1.8M & 0.3M & 4.1M \\
        \bottomrule
    \end{tabular}}
    \caption{Statistics of all datasets. Euro.: \emph{europarl v7}; CC.: \emph{commoncrawl}; NC.: \emph{news commentary v12}.}
    \label{tab:dataset}
\end{table}

\begin{table*}[t]
    \centering
    \resizebox{\textwidth}{!}{
    \begin{tabular}{l|cc|cccccc|cc}
        \toprule
         \multirow{2}{*}{\textbf{Models}} & \multicolumn{2}{c|}{\textbf{External Data}} & \multicolumn{2}{c}{\textbf{En$\rightarrow$De}} & \multicolumn{2}{c}{\textbf{En$\rightarrow$Fr}} & \multicolumn{2}{c}{\textbf{En$\rightarrow$Es}} & \multicolumn{2}{c}{\textbf{Average}} \\
         & \textbf{Audio} & \textbf{Target} & \textbf{BLEU} & \textbf{$\Delta$} & \textbf{BLEU} & \textbf{$\Delta$} & \textbf{BLEU} & \textbf{$\Delta$} & \textbf{BLEU} & \textbf{$\Delta$} \\ 
         \midrule
         \multicolumn{11}{c}{\emph{Previous ST baselines (No transcripts used)}} \\
         \midrule
         \textsc{Revisit-ST} \citep{revisit-st} & \texttimes & \texttimes & 23.0 &      & 33.5 &      & 28.0 &      & 28.2 &      \\
         \textsc{W-Transf.} \citep{ye2021end}  & \checkmark & \texttimes & 23.6 &      & 34.6 &      & 28.4 &      & 28.9 &      \\
         \midrule
         \multicolumn{11}{c}{\emph{Our implementations}} \\
         \midrule
         \textbf{\textsc{Hu-Transformer}}  & \checkmark & \texttimes & 24.3 &      & 34.9 &      & 28.7 &      & 29.3 &      \\
         \textbf{\textsc{BT4ST}}      & \checkmark & \checkmark & \textbf{26.6*} & \textit{\textbf{+2.3}} & \textbf{36.9*} & \textit{\textbf{+2.0}} & \textbf{31.2*} & \textit{\textbf{+2.5}} & \textbf{31.6} & \textit{\textbf{+2.3}} \\
        \bottomrule
    \end{tabular}}
    \caption{BLEU scores on MuST-C En$\rightarrow$De, En$\rightarrow$Fr, and En$\rightarrow$Es \texttt{tst-COMMON} set. * means the improvements over \textbf{\textsc{Hu-Transformer}} are statistically significant ($p<0.01$).}
    \label{tab:main}
\end{table*}

\paragraph{Unit-to-speech Model}
For the \emph{unit-to-speech} model, the configurations of the generator and discriminator are the same as \citet{polyak21_interspeech}. The 1D-convolutional layers in the duration predictor are set to kernel size 3, padding 1, and hidden dimension 256. We use the pre-trained \emph{d-vector} model\footnote{\url{https://github.com/yistLin/dvector}} as the speaker encoder to extract the 256-dimensional speaker embedding. We train the model up to 100k steps with a batch size of 128.

\paragraph{ST Model}
The ST model contains three stacked modules. We use the pre-trained HuBERT-Base model as the acoustic encoder, which takes the raw 16-bit 16kHz audio wave as input. The length adaptor comprises two 1D-convolutional layers with kernel size 5, stride size 2, padding 2, and hidden dimension 1024. The translation model follows Transformer-Base architecture, containing 6 encoder layers and 6 decoder layers. Each layer comprises 512 hidden states, 8 attention heads, and 2048 feed-forward hidden states. The dropout is 0.1, and the label smoothing is 0.1. We refer to this architecture as \textbf{\textsc{Hu-Transformer}}. We learn a vocabulary of size 8k from the target texts in the MuST-C dataset to segment the target texts into subwords for training the ST model.

To augment the ST model with back translation, we first select the top $\rho=75\%$ of synthetic data based on BLEU scores. The target texts in the synthetic data are segmented into subwords using the same vocabulary. We use Adam optimizer with 4k warm-up steps to pre-train the model on synthetic data up to 300k steps, and fine-tune the model on MuST-C up to 20 epochs. During both pre-training and fine-tuning, each batch contains at most 16M audio frames, and the maximum learning rate is 1e-4. During inference, we average the checkpoints of the last 10 epochs for evaluation. We use beam search with a beam size of 8. SacreBLEU\footnote{\url{https://github.com/mjpost/sacrebleu}}~\citep{post-2018-call} is used to compute case-sensitive detokenized BLEU scores and the statistical significance of translation results with paired bootstrap resampling\footnote{sacreBLEU signature: nrefs:1 | bs:1000 | seed:12345 | case:mixed | eff:no | tok:13a | smooth:exp | version:2.0.0}~\citep{koehn-2004-statistical}. The length penalty is set to 1.2, 1.8, and 0.6 for En$\rightarrow$De, En$\rightarrow$Fr, and En$\rightarrow$Es, respectively. We implement our model with \emph{fairseq}\footnote{\url{https://github.com/pytorch/fairseq}}~\citep{ott2019fairseq}. All models are trained on 4 Nvidia Tesla V100 GPUs. 

\begin{figure}[t]
    \centering
    \includegraphics[width=\linewidth]{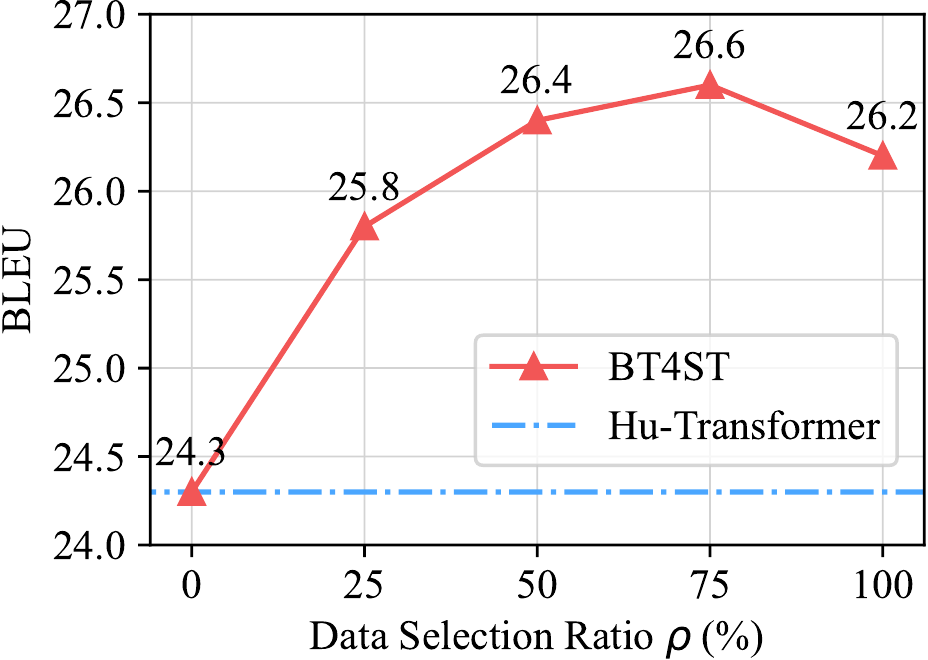}
    \caption{BLEU scores on MuST-C En$\rightarrow$De \texttt{tst-COMMON} set with different data selection ratio $\rho$.}
    \label{fig:data_selection}
\end{figure}

\paragraph{Baseline Systems}
We include two baseline systems: \textsc{Revisit-ST}~\citep{revisit-st} and \textsc{W-Transf.}~\citep{ye2021end} for comparison. \textsc{Revisit-ST} is a carefully designed ST baseline including several techniques like parameterized distance penalty (PDP) and CTC-based regularization. \textsc{W-Transf.} is a stronger ST baseline with a pre-trained acoustic model, which combines Wav2vec 2.0~\citep{baevski2020wav2vec} and Transformer. Both of them are only trained on speech-translation pairs without using any transcripts. Our \textbf{\textsc{Hu-Transformer}} is also a strong baseline model trained on speech-translation pairs from MuST-C, and we examine our proposed \textbf{\textsc{BT4ST}} on top of this by adding synthetic ST data.

\section{Results and Analysis}

\subsection{Results on MuST-C Dataset}

Table \ref{tab:main} shows the results on MuST-C En$\rightarrow$De, En$\rightarrow$Fr, and En$\rightarrow$Es \texttt{tst-COMMON} set. First, we observe that our \textbf{\textsc{Hu-Transformer}} is a strong baseline compared with previous baselines. Second, by synthesizing pseudo ST data with our proposed \textbf{\textsc{BT4ST}} and using it to pre-train the model, we achieve an average boost of 2.3 BLEU in three translation directions. It demonstrates that our approach can effectively utilize external monolingual target data to improve the performance of ST.

\subsection{Impact of Data Selection Ratio}

Although previous work~\citep{edunov-etal-2018-understanding} found that noisy synthetic data can benefit training, we argue that the extremely low-quality data also hurt performance. As described in Section \ref{sec:data_select}, we select top $\rho\cdot M$ synthetic data based on BLEU scores. We constrain $\rho$ in $[0\%, 25\%, 50\%, 75\%, 100\%]$ for experiments, and the results on MuST-C En$\rightarrow$De \texttt{tst-COMMON} set are shown in Figure \ref{fig:data_selection}. We find that filtering out the last $25\%$ of samples (\emph{i.e.}, $\rho=75\%$) gives a 0.4 BLEU improvement (26.2$\rightarrow$26.6) and performs best. We use $\rho=75\%$ for all experiments.

\subsection{Impact of Unit Generation Methods}
\begin{table}[t]
    \centering
    \begin{tabular}{c|c}
        \toprule
        \textbf{Methods} & \textbf{BLEU} \\
        \midrule
         Beam search & 26.6 \\
         Greedy search & 26.2 \\
         Top-10 sampling & 26.4 \\
         Sampling & 26.1 \\
        \bottomrule
    \end{tabular}
    \caption{BLEU scores on MuST-C En$\rightarrow$De \texttt{tst-COMMON} set with different unit generation methods.}
    \label{tab:generation}
\end{table}

The generation method of the \emph{target-to-unit} model also influences the performance of back translation, as it determines the quality and diversity of synthetic data. We conduct experiments with beam search, greedy search, top-10 sampling\footnote{Top-10 sampling refers to sampling over the ten most likely words.}, and sampling. As shown in Table \ref{tab:generation}, beam search performs best over all methods on MuST-C En$\rightarrow$De \texttt{tst-COMMON} set. We consider two reasons for this. First, our ST dataset is actually a low-resource setting, containing only less than 300k parallel data\footnote{\citet{edunov-etal-2018-understanding} find that beam search is more effective than sampling in low-resource settings, while the opposite is true for high-resource settings.}. Second, target-to-unit generation is a more difficult task compared to text translation. Therefore, beam search is more effective as it generates high probability outputs, while sampling from the model distribution may produce harmful low-quality data. We use beam search for all experiments.

\subsection{Single-speaker vs. Multi-speaker Synthetic Data}
\begin{table}[t]
    \centering
    \begin{tabular}{c|c}
        \toprule
        \textbf{Synthetic Data Types} & \textbf{BLEU} \\
        \midrule
         Multi-speaker & 26.6 ($\pm$ 0.1) \\
         Single-speaker & 26.4 \\
        \bottomrule
    \end{tabular}
    \caption{BLEU scores on MuST-C En$\rightarrow$De \texttt{tst-COMMON} set with different types of pseudo data. For multi-speaker data, we report the mean value and standard variance of BLEU scores derived from 3 independent experiments with different random seeds.}
    \label{tab:speaker}
\end{table}

As described in Section \ref{sec:unit2speech}, we provide speaker embedding as input during speech synthesis, which allows us to synthesize pseudo ST datasets containing multiple speakers. To examine whether speaker diversity benefits back translation, we synthesize both multi-speaker pseudo data and single-speaker pseudo data for experiments. For multi-speaker data, we randomly select a speaker embedding from the training set for each sample. For single-speaker data, we use the average speaker embedding on the training set for all samples. As shown in Table \ref{tab:speaker}, we find that the multi-speaker data is slightly better than the single-speaker data, probably because it is closer to the real ST dataset.

\subsection{Results with Different Amounts of Monolingual/Parallel Data}

\begin{figure}[t]
    \centering
    \includegraphics[width=\linewidth]{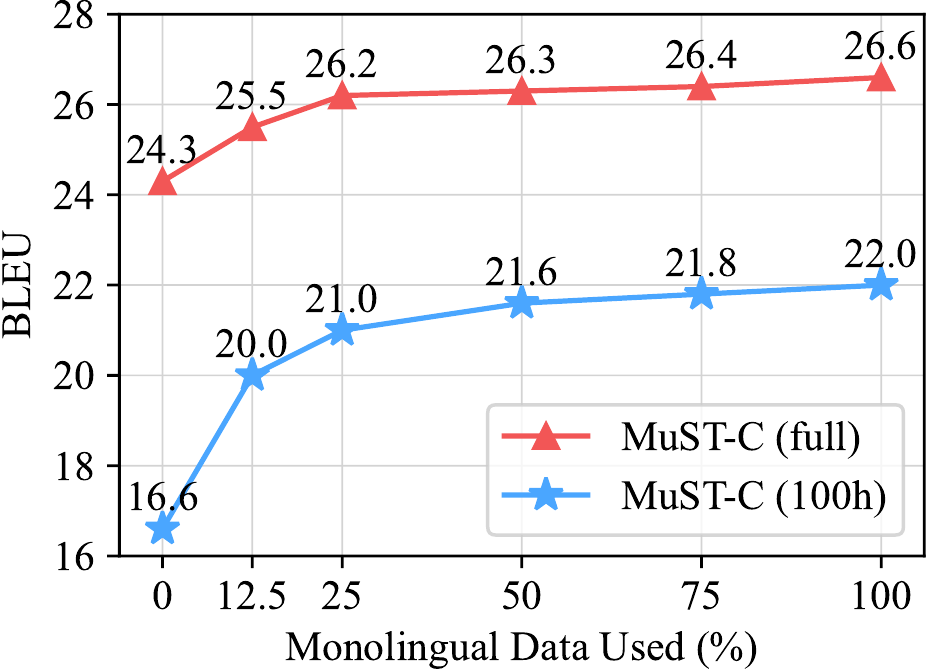}
    \caption{BLEU scores on MuST-C En$\rightarrow$De \texttt{tst-COMMON} set with different amount of monolingual and parallel data.}
    \label{fig:data_size}
\end{figure}

In this section, we investigate the results under different amounts of monolingual and parallel data. First, we randomly sample 100 hours of ST data (corresponding to about 58K speech-translation pairs) from MuST-C En$\rightarrow$De \texttt{train} set to simulate the low-resource setting. We then vary the amount of monolingual data used for back translation. As shown in Figure \ref{fig:data_size}, we observe that (i) the BLEU score keeps increasing with the number of monolingual data regardless of the size of parallel data, and (ii) our approach is particularly effective in the low-resource setting. With only 12.5\% monolingual data (about 0.6M), we observe a 3.4 BLEU improvement compared with the baseline. When all 4.6M monolingual data is used, we achieve a more significant boost of 5.4 BLEU.

\subsection{Diverse BT4ST and Model Ensemble}
\begin{table*}[t]
    \centering
    \resizebox{\textwidth}{!}{
    \begin{tabular}{l|c|cccccc|cc}
        \toprule
         \multirow{2}{*}{\textbf{Models}} & \multirow{2}{*}{\textbf{\#Models}} & \multicolumn{2}{c}{\textbf{En$\rightarrow$De}} & \multicolumn{2}{c}{\textbf{En$\rightarrow$Fr}} & \multicolumn{2}{c}{\textbf{En$\rightarrow$Es}} & \multicolumn{2}{c}{\textbf{Average}} \\
         & & \textbf{BLEU} & \textbf{$\Delta$} & \textbf{BLEU} & \textbf{$\Delta$} & \textbf{BLEU} & \textbf{$\Delta$} & \textbf{BLEU} & \textbf{$\Delta$} \\ 
         \midrule
         \textbf{\textsc{Hu-Transformer}}  & 1 & 24.3 &      & 34.9 &      & 28.7 &      & 29.3 &      \\
         \textbf{\textsc{BT4ST}}     & 1 & 26.6 & \textit{+2.3} & 36.9 & \textit{+2.0} & 31.2 & \textit{+2.5} & 31.6 & \textit{+2.3} \\
         \midrule
         \multirow{4}{*}{\textbf{\textit{Diverse} \textsc{BT4ST} + Ensemble}}     & 2 & 27.5 & \textit{+3.2} & 38.3 & \textit{+3.4} & 32.3 & \textit{+3.6} & 32.7 & \textit{+3.4} \\
              & 3 & 27.9 & \textit{+3.6} & 38.8 & \textit{+3.9} & 32.4 & \textit{+3.7} & 33.0 & \textit{+3.7} \\
              & 4 & \textbf{28.1} & \textbf{\textit{+3.8}} & \textbf{39.0} & \textbf{\textit{+4.1}} & 32.6 & \textit{+3.9} & 33.2 & \textit{+3.9} \\
              & 5 & 28.0 & \textit{+3.7} & \textbf{39.0} & \textbf{\textit{+4.1}} & \textbf{32.8} & \textbf{\textit{+4.1}} & \textbf{33.3} & \textbf{\textit{+4.0}} \\
        \bottomrule
    \end{tabular}}
    \caption{BLEU scores on MuST-C En$\rightarrow$De, En$\rightarrow$Fr, and En$\rightarrow$Es \texttt{tst-COMMON} set with model ensemble.}
    \label{tab:ensemble}
\end{table*}

Model ensemble is a widely-used technique in state-of-the-art MT systems~\citep{barrault-etal-2020-findings, akhbardeh-etal-2021-findings}, which can combine different single models (\emph{e.g.}, models trained on different data) to achieve stronger performance. To synthesize multiple diverse pseudo datasets from a single monolingual corpus, we introduce a simple \textbf{\textit{Diverse} \textsc{BT4ST}} method. For the \emph{target-to-unit} model, we activate the dropout modules during beam search decoding. For the \emph{unit-to-speech} model, we randomly select the speaker embedding as above. By setting different random seeds, we can generate source speeches with different content and different speakers from the target text. In this way, we obtain multiple different pseudo datasets and train several models individually. We then combine these models by computing the token-level average log probability during decoding. As shown in Table \ref{tab:ensemble}, model ensemble can significantly boost performance. We achieve an average boost of 4.0 BLEU in three directions by ensembling five models.

\subsection{Performance of the Target-to-unit Model}
In this section, we report the performance of our target-to-unit models. We evaluate the performance with two metrics: Unit-BLEU and ASR-BLEU. Unit-BLEU is the BLEU score calculated on the reduced discrete unit sequence. ASR-BLEU is the BLEU score calculated on the transcribed text with the open-source ASR-BLEU toolkit\footnote{\url{https://github.com/facebookresearch/fairseq/tree/ust/examples/speech_to_speech/asr_bleu}}. As shown in Table \ref{tab:unit_asr_bleu}, our target-to-unit models achieve promising results on MuST-C De$\rightarrow$En, Fr$\rightarrow$En, and Es$\rightarrow$En \texttt{tst-COMMON} set, indicating that our proposed method can synthesize reasonable ST data.

\begin{table}[t]
    \centering
    \begin{tabular}{c|ccc}
        \toprule
        \textbf{Metrics} & \textbf{De$\rightarrow$En} & \textbf{Fr$\rightarrow$En} & \textbf{Es$\rightarrow$En} \\
        \midrule
        \textbf{Unit-BLEU} & 26.5 & 27.9 & 27.2 \\
        \textbf{ASR-BLEU}  & 19.0 & 24.5 & 20.8 \\
        \bottomrule
    \end{tabular}
    \caption{Unit-BLEU and ASR-BLEU scores of the target-to-unit models.}
    \label{tab:unit_asr_bleu}
\end{table}

\section{Related Work}
\paragraph{End-to-end ST}
End-to-end ST is theoretically attractive due to its advantages in alleviating error propagation and reducing latency, but it also faces many challenges because of data scarcity. Therefore, researchers often leverage \emph{source transcripts} to help train with auxiliary tasks. Plenty of existing work first pre-train the model with the ASR task~\citep{bansal2019pre, stoian2020analyzing, wang-etal-2020-curriculum}, MT task~\citep{han-etal-2021-learning, fang-etal-2022-stemm, ye-etal-2022-cross, fang-and-feng-2023-understanding, zhou-etal-2023-cmot}, or both together~\citep{wang2020bridging, alinejad2020effectively, le-etal-2021-lightweight, dong2021consecutive, xu-etal-2021-stacked}, and then fine-tune the model with the ST task, which becomes the \emph{de-facto} paradigm in recent ST studies. \citet{le-etal-2020-dual, taskaware, tang-etal-2021-improving, generalmtl, dong2021listen, ye2021end} adopt multi-task learning to share knowledge among different tasks to improve ST. \citet{slam, mSLAM, Chen2022MAESTROMS, muslam, ao-etal-2022-speecht5, tang-etal-2022-unified, zhang2022speechut} jointly pre-train the model with speech and text data to learn a unified space for both modalities, which achieve competitive results in ST. \citet{jia2019leveraging, lam-etal-2022-sample} synthesize ST data with the help of MT model, TTS model and forced alignment tools. However, all of these studies assume that transcripts are available, which does not hold true for large numbers of unwritten languages in the world. \citet{revisit-st} first challenge this assumption and propose a set of practices to train a better ST model with only speech-translation pairs. In this paper, we extend this line of research and propose a back translation algorithm to utilize large-scale monolingual target data to improve ST without transcripts.

\paragraph{Back Translation}
Back translation for NMT was first proposed by \citet{sennrich-etal-2016-improving} and is widely used in state-of-the-art NMT systems~\citep{akhbardeh-etal-2021-findings}. Since then, many techniques have been proposed to improve BT such as Iterative BT~\citep{hoang-etal-2018-iterative, dou-etal-2020-dynamic}, Tagged BT~\citep{caswell-etal-2019-tagged, marie-etal-2020-tagged}, Tag-less BT~\citep{taglessbt}, MetaBT~\citep{pham2021meta}, HintedBT~\citep{ramnath-etal-2021-hintedbt}, and so on. \citet{edunov-etal-2018-understanding} investigates different generation methods of BT in large-scale settings. \citet{huang-etal-2021-comparison, liu-etal-2021-complementarity-pre} focus on combining back translation with pre-training. \citet{xu-etal-2022-synthetic} combines synthetic data generated by beam search and sampling to better trade off the importance and quality of synthetic data. Despite the success of BT in MT, BT for ST is still a challenging problem. \citet{nguyen2022improving} introduces a pipeline BT method for speech-to-speech translation which cascades an unsupervised MT model and a TTS model. In contrast, our approach can synthesize pseudo ST data from the target-side monolingual corpus without relying on source transcripts.

\paragraph{Discrete Speech Units}
Discrete units, as a self-supervised discrete representation of speech, have proved effective on many tasks, such as spoken language modeling~\citep{lakhotia-etal-2021-generative, kharitonov-etal-2022-text, gat2022robustness, audiolm}, speech-to-speech translation~\citep{lee-etal-2022-direct, lee-etal-2022-textless, Popuri2022EnhancedDS, inaguma2022unity, enhokkien, TextlessJiaYe}, speech emotion conversion~\citep{SpeechEmotionConversion}, speech dialogue~\citep{speechdialogue}, speech resynthesis~\citep{polyak21_interspeech}, speaking style conversion~\citep{Maimon2022SpeakingSC}, and so on. In this paper, we achieve back translation for ST by leveraging discrete units and further prove its effectiveness.

\section{Conclusion and Future Work}
In this paper, we develop a back translation algorithm for speech translation, which can synthesize pseudo ST data from monolingual target data without relying on transcripts. We utilize self-supervised discrete units and achieve back translation by cascading a \emph{target-to-unit} model and a \emph{unit-to-speech} model. Experimental results on the MuST-C benchmark demonstrate the superiority of our approach, especially in low-resource settings. 

This work focuses on enhancing ST when the source transcripts are unavailable, which is an essential but under-explored issue. We hope our work will draw more attention to this issue from researchers, which will benefit more real-world unwritten languages. In the future, we are interested in exploring how to combine advanced BT techniques (\emph{e.g.}, Iterative BT) with our approach.

\section*{Limitations}
Our work provides an effective solution to augment ST when source transcripts are unavailable, which could benefit many unwritten languages. However, limited by the publicly available ST datasets, we use English as an unwritten language for experiments, which may slightly differ from real-world unwritten languages. Since we never use transcripts in our approach, we believe our work can shed some light on ST for real-world unwritten languages. We are glad to explore this if there are available datasets in the future.

\section*{Ethics Statement}
Our model is developed and evaluated with publicly available datasets: MuST-C and WMT. The pre-trained models we use, like HuBERT and d-vector models, are open and permitted for research purposes. Our use of the above artifacts is consistent with their intended use since they are widely used in the speech research community. Although our method could help the speech translation of unwritten languages like some dialects, the performance of the ST model still heavily relies on the amount of ST training data. Therefore, the output of the model is not always reliable and it would be better to be assisted by professional human translators in real applications.

\section*{Acknowledgements}
We thank all the anonymous reviewers for their insightful and valuable comments. This work was supported by National Key R\&D Program of China (NO. 2018AAA0102502)

\bibliography{anthology,custom}
\bibliographystyle{acl_natbib}

\appendix
\clearpage


\section{Details about the HiFi-GAN}
\label{appendix:hifigan}

The HiFi-GAN vocoder~\citep{hifi-gan} consists of one generator and two discriminators. Next, we describe the model architecture and training objectives of HiFi-GAN.

\paragraph{Generator}
The generator is a convolutional neural network containing several stacked blocks. Each block comprises a transposed convolution layer followed by a multi-receptive field fusion (MRF) module. The transposed convolution layers upsample the input sequence to match the length of output waveforms. The MRF module includes multiple residual blocks with different receptive fields to model different patterns in parallel. The input to the generator is concatenated unit embedding ${\rm Emb}(\mathbf{z})$ and speaker embedding $\mathbf{e}_{\rm spkr}$.

\paragraph{Discriminator}
The discriminator consists of a Multi-Period Discriminator (MPD) and a Multi-Scale Discriminator (MSD). MPD is a mixture of sub-discriminators operating on equally spaced samples of the input audio. We adopt 5 sub-discriminators and set the space between samples to $[2, 3, 5, 7, 11]$ respectively. MSD is also a mixture of sub-discriminators operating on different input scales: raw audio, $\times 2$ downsampled audio, and $\times 4$ downsampled audio. MPD and MSD can identify different periodic and consecutive patterns in the audio.

\paragraph{Training Objectives}
Using $G$ to denote the generator and $D_j$ to denote each sub-discriminator, we define the adversarial loss of $G$ and $D_j$ as:
\begin{gather}
    \mathcal{L}_{\rm adv}(G; D_j) = \mathbb{E}_\mathbf{x}\left[(1-D_j(\widehat{\mathbf{x}}))^2\right], \\
    \mathcal{L}_{\rm adv}(D_j; G) = \mathbb{E}_\mathbf{x}\left[(1-D_j(\mathbf{x}))^2+D_j(\widehat{\mathbf{x}})^2\right],
\end{gather}
where $\mathbf{x}$ denotes the ground truth audio and $\widehat{\mathbf{x}}=G({\rm Emb}(\mathbf{z}), \mathbf{e}_{\rm spkr})$ denotes the synthetic audio.

Besides, there are two auxiliary training objectives. The first term measures the L1 distance between mel-spectrogram of the ground truth audio and synthetic audio:
\begin{equation}
    \mathcal{L}_{\rm mel}(G) = \mathbb{E}_\mathbf{x}\left[\|\phi(\mathbf{x}) -  \phi(\widehat{\mathbf{x}})\|_1\right],
\end{equation}
where $\phi$ is the function to compute mel-spectrogram of the audio. The second term is a feature-matching loss which measures the difference in discriminator features between ground truth audio and synthetic audio:
\begin{equation}
    \mathcal{L}_{\rm fm}(G;D_j)=\mathbb{E}_\mathbf{x}\left[\sum_{i=1}^{L}\frac{1}{N_i}\|\delta_i(\mathbf{x})-\delta_i(\widehat{\mathbf{x}})\|_1\right],
\end{equation}
where $L$ denotes the number of layers in $D_j$, $\delta_i$ denotes the feature extractor of $i$-th layer, and $N_i$ denotes the number of features in $i$-th layer.

Considering all sub-discriminators, the final training objectives are as follows:
\begin{align}
    \mathcal{L}_G =& \sum_{j=1}^J[\mathcal{L}_{\rm adv}(G;D_j)+\lambda_{\rm fm}\mathcal{L}_{\rm fm}(G;D_j)] \\
    &+ \lambda_{\rm mel}\mathcal{L}_{\rm mel}(G), \\
    \mathcal{L}_D=&\sum_{j=1}^J \mathcal{L}_{\rm adv}(D_j; G),
\end{align}
where $J$ is the number of sub-discriminators. We set $\lambda_{\rm fm}=2$ and $\lambda_{\rm mel}=45$.


\section{Examples of Synthetic ST Data}
\label{sec:case}

To understand our approach more intuitively, we provide some examples of synthetic ST data in this section. Table \ref{tab:case-ende}, \ref{tab:case-enfr}, and \ref{tab:case-enes} show some examples of synthetic En$\rightarrow$De, En$\rightarrow$Fr, and En$\rightarrow$Es ST data, respectively. For each sample, we give the original target text, generated \emph{reduced} discrete units, generated source speech, and the corresponding transcript obtained with a state-of-the-art ASR model Whisper-Large\footnote{\url{https://github.com/openai/whisper}}~\citep{whisper}. We observe that our method can generate reasonable source speech, though it may contain minor errors or duplicates. This explains why our method can enhance ST successfully. 

We also provide an example of our proposed \textbf{\emph{Diverse} \textsc{BT4ST}} method in Table \ref{tab:case-diverse}, which generates multiple diverse source speeches from only one target text. We observe that the four outputs of our model are all reasonable and differ from each other in some way, which confirms that \textbf{\emph{Diverse} \textsc{BT4ST}} is a simple and effective method to generate diverse pseudo data.

We provide the corresponding audio files of the above samples at \url{https://bt4st.github.io/}.

\begin{table*}[t]
    \centering
    \resizebox{\textwidth}{!}{
    \begin{tabular}{l|p{23cm}}
    \toprule
    \multicolumn{2}{c}{\textbf{\textsc{BT4ST}: German text$\rightarrow$English speech}} \\
    \midrule
    \multicolumn{2}{c}{\textbf{Case 1}} \\
    \midrule
    \multirow{2}{*}{\textbf{Target (De)}} & Die Art und Weise, wie wir diese Steuern zahlen, wird sich verändern. \\ 
    & (\emph{There is going to be a change in how we pay these taxes.}) \\
    \midrule
    \multirow{2}{*}{\textbf{Unit}} & 71 82 73 70 14 76 94 32 64 60 70 14 76 11 64 74 27 47 59 33 94 32 64 1 66 82 11 45 64 65 6 15 92 57 31 59 33 91 43 74 2 89 6 15 7 23 53 62 29 28 60 70 14 76 53 97 19 65 74 27 21 95 59 87 94 32 64 81 57 21 95 41 20 \\
    \midrule
    \textbf{Speech (En)} & \makecell[c]{\includegraphics[width=0.99\linewidth]{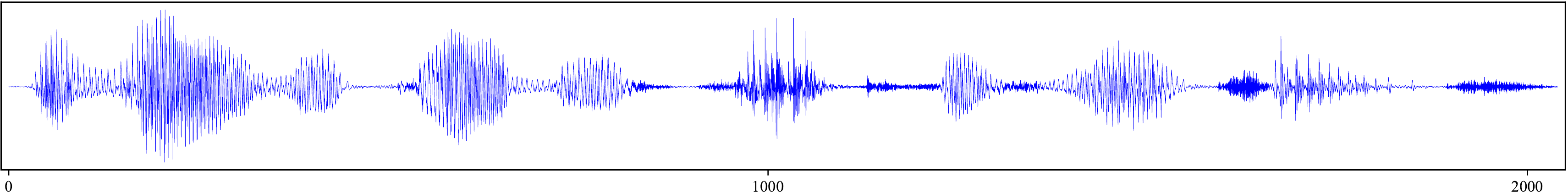}} \\
    \midrule
    \textbf{ASR Output} & The way we pay these taxes will change. \\
    \midrule
    \multicolumn{2}{c}{\textbf{Case 2}} \\
    \midrule
    \multirow{2}{*}{\textbf{Target (De)}} & Auch diese Frage soll letztlich Aufschluss darüber geben, welche Voraussetzungen es für die Entstehung von Leben gibt. \\
    & (\emph{This question should also provide information regarding the preconditions for the origins of life.}) \\
    \midrule
    \multirow{5}{*}{\textbf{Unit}} & 71 86 38 44 80 26 87 91 43 74 2 78 33 14 76 68 9 43 6 95 92 21 95 23 42 44 80 81 83 84 57 96 55 39 67 54 57 93 86 53 62 29 28 37 24 51 19 74 2 31 59 23 16 50 87 53 9 74 2 90 35 11 64 1 66 47 11 45 64 74 27 89 59 23 44 80 18 27 78 33 90 35 69 65 29 95 23 42 80 81 83 84 57 96 55 39 67 54 57 93 86 68 73 16 66 47 87 91 17 19 70 2 70 14 68 9 74 2 27 89 59 23 44 80 18 66 31 53 65 6 95 23 42 44 80 18 6 15 49 41 84 57 96 55 67 54 57 93 82 87 0 30 37 24 61 46 79 81 83 2 96 55 67 54 57 93 3 48 46 30 99 82 11 64 53 73 16 50 52 30 1 21 95 23 44 80 18 6 15 49 7 23 73 16 66 77 90 35 24 13 58 32 81 65 3 41 20 \\
    \midrule
    \textbf{Speech (En)} & \makecell[c]{\includegraphics[width=0.99\linewidth]{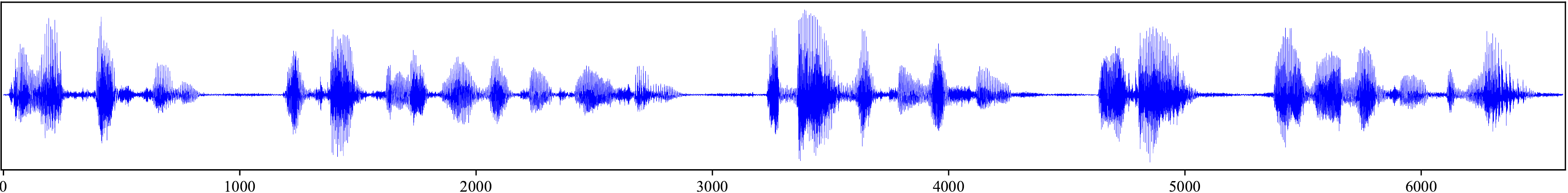}} \\
    \midrule
    \textbf{ASR Output} & And that question is ultimately the conclusion about what conditions there are for the emergence of life. \\
    \midrule
    \multicolumn{2}{c}{\textbf{Case 3}} \\
    \midrule
    \multirow{4}{*}{\textbf{Target (De)}} & Doch die von ihr getesteten Geräte sind auch für die Straßenplaner interessant, denn sie arbeiten nicht mit GPS und liefern nur begrenzte Informationen, die regelmäßig per Modem hochgeladen werden. \\
    & (\emph{But the devices it is testing appeal to highway planners because they don't use GPS and deliver a limited amount of information, uploaded periodically by modem.}) \\
    \midrule
    \multirow{6}{*}{\textbf{Unit}} & 71 47 76 9 74 2 82 11 45 64 29 28 92 31 23 73 16 77 24 13 58 32 65 6 15 7 23 62 29 28 92 82 87 94 32 64 74 27 31 59 33 91 43 6 49 92 31 23 62 1 85 5 30 37 51 19 65 6 49 7 87 97 19 37 86 53 44 80 18 21 95 52 25 62 6 49 92 31 87 42 88 81 83 84 57 96 55 39 67 54 57 93 3 52 30 99 82 62 6 49 92 21 52 25 45 64 74 2 27 47 59 33 90 35 13 91 38 44 80 85 5 79 29 6 49 41 84 57 96 55 39 67 54 57 93 47 11 45 64 74 27 89 59 33 68 9 29 28 92 82 87 94 32 64 1 66 31 87 97 51 19 2 66 31 23 69 70 14 46 30 74 2 78 14 76 62 1 66 21 95 45 64 74 27 47 59 45 64 87 91 43 6 15 49 41 84 57 96 55 39 67 54 57 86 38 44 80 18 66 31 23 73 90 35 53 16 50 77 53 1 85 53 1 85 53 44 80 18 65 3 52 30 16 50 87 94 32 64 65 95 23 42 80 81 83 84 57 96 55 67 54 57 93 82 87 9 85 5 30 70 52 25 9 32 1 66 89 98 53 90 35 5 30 90 35 11 64 37 68 43 74 2 47 90 35 97 1 85 23 62 1 66 27 47 24 13 58 16 50 24 61 9 85 42 16 81 20 \\
    \midrule
    \textbf{Speech (En)} & \makecell[c]{\includegraphics[width=0.99\linewidth]{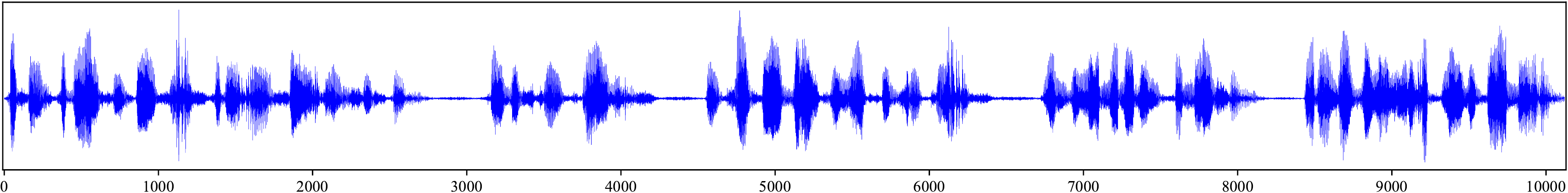}} \\
    \midrule
    \multirow{2}{*}{\textbf{ASR Output}} & But these devices they tested are also interesting for the street planners because they don't do work with GPS and the limited information that are regularly uploaded by modem. \\
    \midrule
    \multicolumn{2}{c}{\textbf{Case 4}} \\
    \midrule
    \multirow{4}{*}{\textbf{Target (De)}} & Diese Kartelle sind dumm, wenn sie meinen, sie könnten sich unter dem Radar hinweg durchgraben, sagte die Generalstaatsanwältin des US Distrikts Southern California, Laura Duffy, bei einer Pressekonferenz vor einem Lagerhaus in San Diego, wo das eine Ende des Tunnels entdeckt worden war. \\
    & (\emph{These cartels are stupid if they think that they can dig through under the radar, said the US Attorney for the District of Southern California, Laura Diffy, at a press conference held in front of a warehouse in San Diego, where the end of the tunnel was discovered.}) \\
    \midrule
    \multirow{9}{*}{\textbf{Unit}} & 71 82 11 45 64 29 28 92 89 23 62 74 27 31 59 33 17 51 19 29 28 5 30 65 6 49 92 31 87 69 74 27 47 76 53 1 85 53 65 3 77 98 69 65 3 77 87 53 88 74 2 89 98 69 74 27 89 23 62 1 66 31 87 53 32 1 66 2 3 59 52 25 69 81 84 96 55 39 67 54 57 93 86 68 44 80 18 85 5 30 99 82 73 70 52 25 94 32 64 1 85 87 24 61 46 79 81 83 84 57 96 55 39 67 54 57 93 82 62 1 66 21 95 87 38 44 80 60 52 25 19 74 27 47 33 52 24 61 43 6 15 49 7 23 62 74 27 31 59 69 1 85 5 30 25 73 16 99 82 11 98 69 14 76 87 9 43 6 15 7 87 94 32 64 1 66 31 53 62 6 49 92 21 52 25 42 74 2 31 41 84 57 96 55 39 67 54 57 93 6 15 7 87 9 1 66 6 15 49 7 87 68 99 82 5 30 44 80 18 27 89 59 33 91 17 19 73 65 3 48 46 30 44 80 98 53 42 81 83 84 57 96 55 39 67 54 57 93 90 35 48 46 30 25 73 1 66 31 87 68 43 3 77 11 64 81 83 84 57 96 55 39 67 54 57 93 86 91 9 85 73 70 14 76 94 0 30 75 91 17 43 6 15 49 92 89 33 24 61 68 44 80 18 65 3 52 25 42 44 80 18 2 6 15 49 41 84 57 96 55 39 67 54 57 93 86 53 44 80 18 6 15 49 7 87 38 44 80 18 85 11 64 53 94 32 1 66 89 87 97 19 81 83 84 57 96 55 39 67 54 57 93 70 14 76 0 30 70 14 68 44 80 85 87 38 44 80 18 85 23 73 16 99 82 73 62 74 27 31 59 33 68 44 80 85 19 70 14 76 62 29 28 92 31 23 62 6 49 92 89 87 68 16 77 5 79 1 85 10 83 20 \\
    \midrule
    \textbf{Speech (En)} & \makecell[c]{\includegraphics[width=0.99\linewidth]{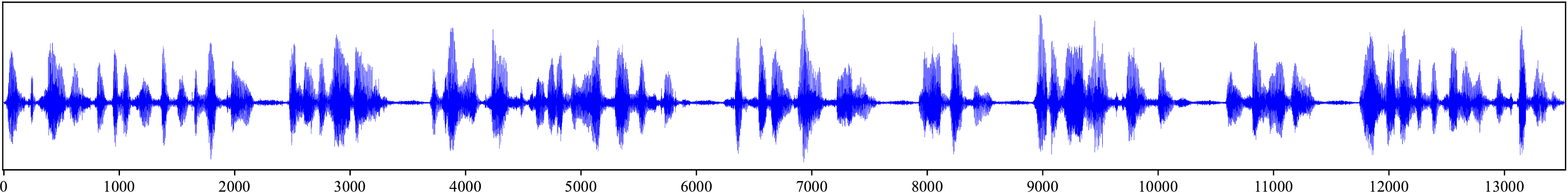}} \\
    \midrule
    \multirow{2}{*}{\textbf{ASR Output}} & These cattles are stupid if you think you could dig through under the radar, the general prostitutor of the USA district, said Southern California, Laura Duffy, at a warehouse conference in San Diego, where one end of the tunnel was discovered. \\
    \bottomrule
    \end{tabular}}
    \caption{Examples of synthetic En$\rightarrow$De ST data.}
    \label{tab:case-ende}
\end{table*}

\begin{table*}[t]
    \centering
    \resizebox{\textwidth}{!}{
    \begin{tabular}{l|p{23cm}}
    \toprule
    \multicolumn{2}{c}{\textbf{\textsc{BT4ST}: French text$\rightarrow$English speech}} \\
    \midrule
    \multicolumn{2}{c}{\textbf{Case 1}} \\
    \midrule
    \multirow{2}{*}{\textbf{Target (Fr)}} & Ces automobilistes paieront bientôt à État des frais kilométriques au lieu des taxes sur essence. \\ 
    & (\emph{Those drivers will soon pay the mileage fees instead of gas taxes to the state.}) \\
    \midrule
    \multirow{3}{*}{\textbf{Unit}} & 71 82 11 45 64 29 28 37 24 61 9 85 73 16 50 14 19 16 66 47 11 53 90 35 62 6 15 49 7 69 44 80 60 70 14 19 74 27 57 47 59 33 94 32 64 65 6 15 49 92 57 31 87 94 32 64 74 2 66 50 24 13 58 17 19 65 3 77 11 45 64 81 84 57 96 55 39 67 54 57 86 53 44 80 18 6 15 49 92 57 31 87 9 85 73 16 77 66 27 89 87 91 43 6 15 7 23 19 90 35 11 64 44 80 18 27 31 59 33 91 43 2 89 6 15 7 11 64 81 29 6 15 41 20 \\
    \midrule
    \textbf{Speech (En)} & \makecell[c]{\includegraphics[width=0.99\linewidth]{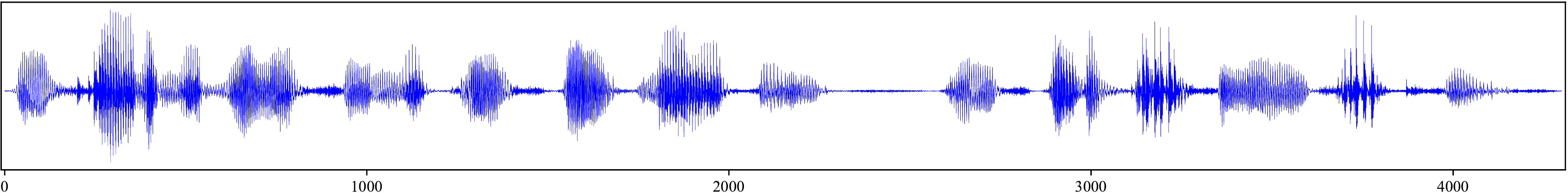}} \\
    \midrule
    \textbf{ASR Output} & These automobilists soon will pay state mile fee instead of gasoline taxis. \\
    \midrule
    \multicolumn{2}{c}{\textbf{Case 2}} \\
    \midrule
    \multirow{2}{*}{\textbf{Target (Fr)}} & Un scandale sur la présence de viande de cheval dans des plats cuisinés a éclaté en Europe au début de année, à la suite de tests effectués en Irlande. \\
    & (\emph{A scandal on the presence of horse meat in prepared meals had broken out in Europe at the beginning of the year, following tests carried out in Ireland.}) \\
    \midrule
    \multirow{5}{*}{\textbf{Unit}} & 71 86 62 6 15 49 92 89 87 91 38 44 80 18 85 19 90 35 73 16 99 82 73 74 27 47 52 25 9 29 28 23 44 80 18 6 15 7 23 73 16 77 75 33 48 51 19 65 6 49 92 50 11 45 64 2 27 31 41 84 57 96 55 39 67 54 57 93 86 53 44 80 18 27 89 59 53 74 2 21 95 23 44 80 18 66 31 53 65 6 95 23 53 62 29 28 92 27 47 52 25 97 65 74 27 89 59 33 91 17 43 2 31 23 53 44 80 18 98 69 48 46 30 25 42 16 74 2 47 41 83 84 57 96 55 39 67 54 57 93 86 53 44 80 82 11 64 0 87 9 90 35 11 64 87 94 32 64 1 21 95 23 73 16 77 98 45 64 0 42 79 81 83 84 57 96 55 39 67 54 57 93 86 91 43 3 2 31 23 73 60 70 14 76 62 29 28 92 27 47 52 25 97 65 74 27 89 23 44 80 18 27 31 59 87 91 43 6 15 49 92 31 23 53 1 85 53 44 80 26 24 13 58 90 35 42 80 18 81 31 41 20 \\
    \midrule
    \textbf{Speech (En)} & \makecell[c]{\includegraphics[width=0.99\linewidth]{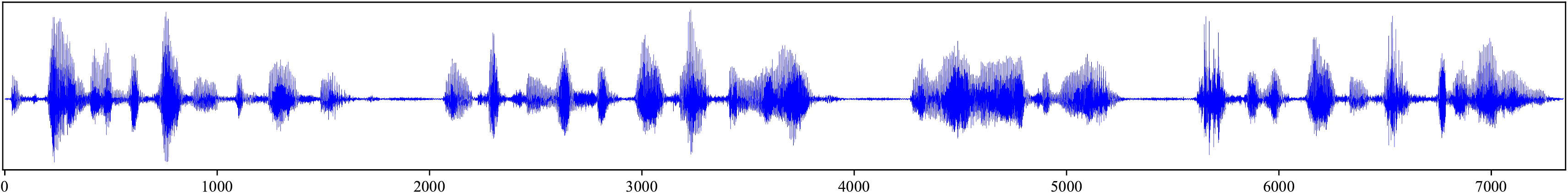}} \\
    \midrule
    \textbf{ASR Output} & A scandal of the presence of horse meat in kitchen dishes pro-cut in Europe in the early age of year after it was pro-contested in Ireland. \\
    \midrule
    \multicolumn{2}{c}{\textbf{Case 3}} \\
    \midrule
    \multirow{3}{*}{\textbf{Target (Fr)}} & La seule autre compagnie imposant de tels frais est la compagnie hongroise Wizz Air, a déclaré le consultant auprès de compagnies aériennes Jay Sorensen, qui suit de près les frais en supplément. \\
    & (\emph{The only other airline with such a fee is Hungary Wizz Air, said airline consultant Jay Sorensen, who closely tracks add-on fees.}) \\
    \midrule
    \multirow{6}{*}{\textbf{Unit}} & 71 82 11 45 64 37 51 90 35 11 64 37 68 99 82 5 30 65 3 52 25 11 45 64 29 28 92 27 89 59 33 68 16 74 2 47 23 44 80 85 11 64 65 6 15 7 87 68 9 74 21 95 23 62 29 28 49 3 77 11 45 64 29 28 41 84 57 96 55 39 67 54 57 93 86 53 62 29 28 92 82 11 45 64 37 86 94 0 30 75 33 68 44 80 18 66 89 98 87 0 30 25 11 32 53 44 80 60 70 14 76 53 62 29 28 60 70 14 76 53 62 29 28 75 33 68 44 80 18 66 89 98 87 0 30 25 11 32 53 44 80 18 27 89 59 33 68 16 18 2 47 23 42 44 80 85 11 64 81 83 84 57 96 55 39 67 54 57 93 6 15 7 87 9 1 66 82 73 74 27 89 59 23 44 80 18 6 15 49 7 24 51 19 74 2 31 23 53 88 18 27 31 59 23 69 1 66 21 95 87 94 32 64 65 6 15 49 7 60 48 46 30 25 44 80 18 6 15 7 23 44 80 37 94 0 30 90 35 24 13 58 32 44 80 18 29 28 15 41 84 57 96 55 39 67 54 57 93 3 24 61 51 19 90 35 97 14 76 88 18 74 27 78 33 90 35 97 65 6 49 92 31 23 99 82 73 74 27 78 33 24 61 43 6 15 49 92 31 41 20 \\
    \midrule
    \textbf{Speech (En)} & \makecell[c]{\includegraphics[width=0.99\linewidth]{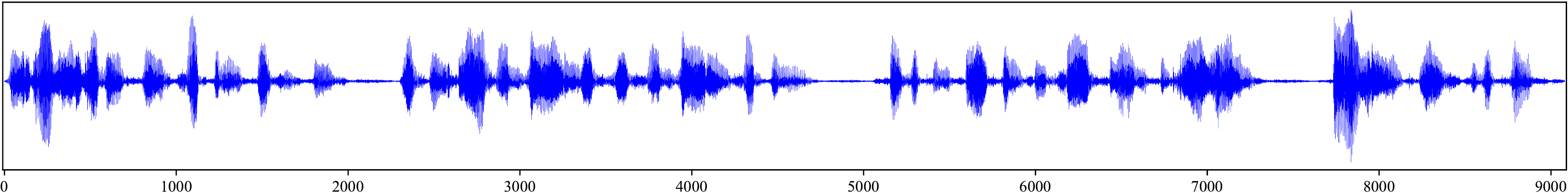}} \\
    \midrule
    \multirow{2}{*}{\textbf{ASR Output}} & The only other freeze companies such as fees is the Air Hungarian with Hungarian company said the consulting to Jay Sorensen Airlines following close to the cost. \\
    \midrule
    \multicolumn{2}{c}{\textbf{Case 4}} \\
    \midrule
    \multirow{6}{*}{\textbf{Target (Fr)}} & Ces dernières accusations se basent entre autres sur des lettres rédigées par avocat des frères Sainte-Croix, Me Émile Perrin, dans les années 1990, mais aussi par les recherches faites dans les archives à ce sujet par le frère Wilson Kennedy, un ancien frère de Sainte-Croix qui a dénoncé publiquement les sévices. \\
    & (\emph{The latter accusations are partly based on letters written by the lawyer of the brothers of the Holy Cross, Mr Emile Perrin QC, in the 1990s, as well as through research carried out in the archives on this subject by Brother Wilson Kennedy, a former brother of the Holy Cross who has publicly denounced the abuses.}) \\
    \midrule
    \multirow{11}{*}{\textbf{Unit}} & 71 86 53 44 80 82 73 90 35 24 61 43 6 15 49 92 31 41 84 96 55 39 67 54 57 93 86 91 43 74 2 89 98 69 29 28 87 94 32 65 6 95 23 42 80 18 29 28 49 41 84 96 55 67 54 57 93 86 24 61 46 30 25 73 16 50 24 68 80 18 90 35 87 9 43 74 2 31 59 23 42 29 28 49 41 84 96 55 67 54 57 93 70 52 25 53 74 2 23 80 18 66 47 24 13 58 9 90 35 48 46 76 58 32 0 42 29 28 49 41 84 96 55 67 54 57 93 86 73 16 99 82 73 75 33 97 90 35 11 64 74 27 78 52 24 61 43 6 49 92 47 52 24 68 99 82 5 42 29 28 49 41 84 96 55 67 54 57 93 86 73 16 50 11 45 64 53 90 35 94 32 64 74 2 6 49 92 50 11 64 81 84 96 55 67 54 57 93 86 53 44 80 82 73 44 80 26 24 13 58 44 18 27 31 59 45 64 80 26 24 13 58 44 18 27 31 59 11 64 29 28 49 41 84 96 55 67 54 57 93 47 76 9 85 60 48 51 19 65 6 49 7 87 97 69 81 84 96 55 67 54 57 93 3 99 60 52 25 69 60 70 52 25 11 45 64 65 6 49 7 87 42 9 74 2 21 95 92 31 87 68 44 80 85 53 44 80 82 11 64 87 24 61 43 74 2 89 59 33 24 13 58 42 16 77 3 41 84 96 55 67 54 57 93 86 73 16 66 47 87 91 17 43 2 66 82 87 91 43 2 31 41 84 96 55 67 54 57 93 47 24 13 58 9 99 82 73 70 14 76 97 19 65 6 49 7 23 44 80 18 27 89 59 33 38 44 80 85 42 62 1 85 11 64 81 84 96 55 67 54 57 93 86 73 3 48 51 16 50 76 73 16 66 47 52 24 68 99 82 5 30 73 16 77 90 35 97 69 65 74 27 78 52 24 61 43 6 49 7 23 42 29 28 49 41 84 96 55 67 54 57 93 82 73 74 2 27 47 33 68 16 66 47 90 35 53 74 2 89 90 35 11 64 1 66 31 23 73 44 80 26 87 91 17 44 18 2 6 49 92 31 41 84 96 55 67 54 57 93 82 62 6 49 7 87 9 16 77 62 6 49 7 42 29 28 49 41 20 \\
    \midrule
    \textbf{Speech (En)} & \makecell[c]{\includegraphics[width=0.99\linewidth]{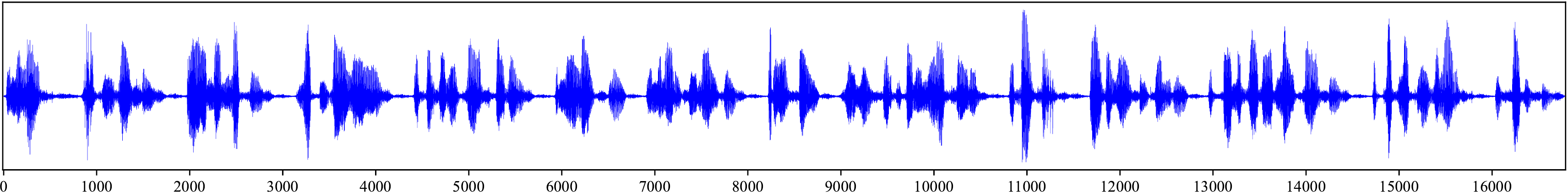}} \\
    \midrule
    \multirow{2}{*}{\textbf{ASR Output}} & In the last. Accusations are among letters written by lawyers of the Holy Cross Brothers ameliates me in the 1990s, but also through research done in the archive about that by the Wilson-Kennedy, a former Brother of Low Crosses that publicly denounced the services. \\
    \bottomrule
    \end{tabular}}
    \caption{Examples of synthetic En$\rightarrow$Fr ST data.}
    \label{tab:case-enfr}
\end{table*}

\begin{table*}[t]
    \centering
    \resizebox{\textwidth}{!}{
    \begin{tabular}{l|p{23cm}}
    \toprule
    \multicolumn{2}{c}{\textbf{\textsc{BT4ST}: Spanish text$\rightarrow$English speech}} \\
    \midrule
    \multicolumn{2}{c}{\textbf{Case 1}} \\
    \midrule
    \multirow{2}{*}{\textbf{Target (Es)}} & En la mayoría de las familias, cada uno desayuna solo. \\ 
    & (\emph{In a majority of families, everyone has breakfast separately.}) \\
    \midrule
    \multirow{2}{*}{\textbf{Unit}} & 71 86 53 44 80 18 50 24 97 65 6 49 92 3 77 87 94 38 17 16 50 90 35 11 64 29 28 49 41 84 57 96 55 67 54 57 40 57 93 86 45 64 74 21 95 60 70 14 68 44 80 18 66 47 52 25 91 43 74 2 3 77 23 62 6 49 92 31 23 73 19 90 35 24 68 97 81 20 \\
    \midrule
    \textbf{Speech (En)} & \makecell[c]{\includegraphics[width=0.99\linewidth]{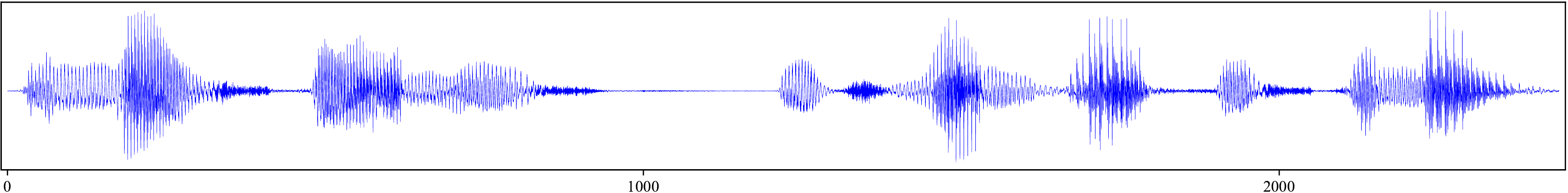}} \\
    \midrule
    \textbf{ASR Output} & In most families, each one breakfast alone. \\
    \midrule
    \multicolumn{2}{c}{\textbf{Case 2}} \\
    \midrule
    \multirow{2}{*}{\textbf{Target (Es)}} & Os sentáis al volante en la costa oeste, en San Francisco, y vuestra misión es llegar los primeros a Nueva York. \\
    & (\emph{You get in the car on the west coast, in San Francisco, and your task is to be the first one to reach New York.}) \\
    \midrule
    \multirow{4}{*}{\textbf{Unit}} & 71 93 98 45 65 6 15 7 87 91 9 74 2 31 87 91 17 68 44 80 85 68 44 80 82 73 70 14 76 45 64 53 17 19 35 68 44 80 82 73 70 14 76 87 91 43 6 15 49 92 57 89 59 33 87 97 65 6 15 49 92 57 31 23 53 44 80 18 6 15 7 87 38 43 16 3 77 23 44 18 6 15 7 23 62 6 15 49 92 57 89 87 97 14 76 44 80 98 5 30 16 50 87 53 65 6 95 23 42 44 80 85 53 62 29 6 15 92 57 31 23 73 74 27 57 89 59 33 68 16 50 77 53 44 80 18 27 31 59 23 44 80 26 11 98 0 5 46 30 74 2 89 6 15 7 53 1 85 11 64 81 83 20 \\
    \midrule
    \textbf{Speech (En)} & \makecell[c]{\includegraphics[width=0.99\linewidth]{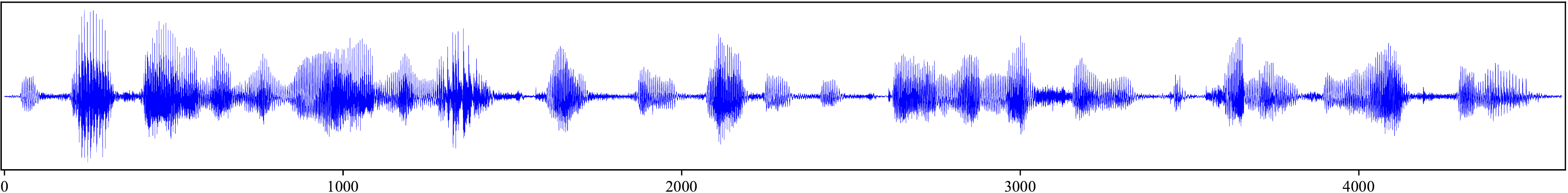}} \\
    \midrule
    \textbf{ASR Output} & You sat down on the wheel on the west coast in San Francisco and your mission is to come into New York City. \\
    \midrule
    \multicolumn{2}{c}{\textbf{Case 3}} \\
    \midrule
    \multirow{4}{*}{\textbf{Target (Es)}} & En la redacción seguramente tendremos una discusión sobre esto durante varios días, sobre si durante la programación del juego ninguno pensó siquiera un poco o si los autores de verdad nos toman por una panda de bobos. \\
    & (\emph{For several days, our editorial staff kept discussing whether when designing the game no one thought for a moment or whether its authors really think we are such morons.}) \\
    \midrule
    \multirow{6}{*}{\textbf{Unit}} & 71 86 53 44 80 60 52 25 11 64 1 66 31 87 91 43 74 2 21 95 23 44 80 60 70 14 76 45 64 60 70 14 76 53 97 19 65 6 95 5 30 90 35 11 64 75 91 9 16 77 23 44 80 37 24 46 30 1 66 89 98 53 16 50 77 42 44 80 85 73 16 66 47 87 91 17 43 74 2 82 53 62 6 49 60 3 52 30 65 6 49 7 87 9 16 77 52 25 19 1 66 31 87 94 32 64 29 28 41 84 57 96 55 39 67 54 57 93 70 14 76 9 99 82 5 30 1 66 31 87 5 30 25 88 18 66 27 47 59 33 90 35 94 32 64 74 27 47 52 25 97 16 66 78 52 25 38 16 50 77 88 18 26 87 68 44 80 85 73 16 99 82 87 42 16 50 81 84 57 96 55 39 67 54 57 93 45 64 16 77 23 44 80 18 6 3 7 24 61 9 85 73 16 66 47 87 53 9 74 2 37 48 46 30 70 14 76 9 99 82 5 30 99 82 11 64 87 91 43 74 2 21 95 60 14 19 37 24 61 43 99 3 82 5 30 29 28 49 41 84 57 96 55 67 54 57 40 57 31 59 94 32 74 2 89 87 68 43 6 49 41 84 96 55 67 54 57 93 3 52 30 73 16 66 47 87 94 38 44 80 18 85 73 16 65 6 49 7 69 81 29 28 49 41 20 \\
    \midrule
    \textbf{Speech (En)} & \makecell[c]{\includegraphics[width=0.99\linewidth]{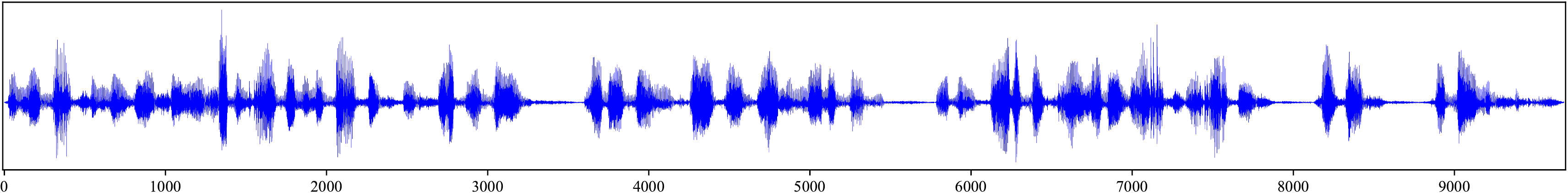}} \\
    \midrule
    \multirow{2}{*}{\textbf{ASR Output}} & In reduction, we will surely have an argument about this for several days, whether during play programming none of them even thought of it, or whether the actual authors take us for a band of zoos. \\
    \midrule
    \multicolumn{2}{c}{\textbf{Case 4}} \\
    \midrule
    \multirow{5}{*}{\textbf{Target (Es)}} & El avión especial, en el que tuvieron que viajar a Podgorica, entre otros los vicepresidentes Dalibor Kucera y Rajchl, no ha salido esta mañana por problemas técnicos en Praga, teniendo que buscarse una alternativa para transportar a parte del comité ejecutivo al lugar donde tendrá lugar el partido de clasificación. \\
    & (\emph{An extra flight that was to take, among others, the Vice Presidents Dalibor Kučera and Rajchl to Podgorica did not leave Prague in the morning due to a technical failure. An alternative was sought for in order to transport part of the executive board to the scene of the barraged rematch.}) \\
    \midrule
    \multirow{11}{*}{\textbf{Unit}} & 71 82 62 6 49 92 47 87 9 43 6 95 23 19 37 86 0 30 74 2 47 59 33 90 35 94 32 64 44 80 60 70 14 46 30 99 82 87 94 32 75 91 9 1 66 31 23 62 74 27 21 59 52 25 91 17 16 77 19 74 27 31 23 73 74 27 47 33 24 61 43 1 66 78 48 46 30 25 42 43 74 2 89 41 83 84 57 96 55 39 67 54 57 93 86 73 16 50 24 68 88 18 37 68 99 82 5 79 29 28 49 41 84 96 55 67 54 57 93 82 73 16 66 77 24 13 58 32 65 6 49 92 47 52 25 9 29 28 23 1 85 53 42 44 80 18 2 6 15 49 41 84 96 55 67 54 57 93 66 31 87 91 17 19 90 35 73 16 66 47 48 46 30 74 2 27 78 33 69 65 74 27 89 59 98 0 30 25 37 86 38 44 80 18 70 52 25 91 43 74 2 21 95 60 48 19 81 83 84 57 96 55 39 67 54 57 93 75 91 9 29 28 92 26 24 61 43 74 2 89 59 33 68 16 50 87 91 17 43 74 2 82 62 6 49 92 50 48 46 30 80 85 42 88 81 83 84 57 96 55 67 54 57 93 3 52 30 16 74 27 47 52 24 61 43 1 66 2 27 31 59 87 9 43 74 2 26 73 74 27 89 78 19 74 27 47 52 24 61 16 66 47 90 35 42 16 50 18 29 28 49 41 84 96 55 39 67 54 57 93 75 91 9 16 77 88 18 27 31 59 23 73 90 35 76 74 2 3 52 30 44 80 26 24 51 19 65 74 27 31 59 33 52 30 44 80 85 53 1 85 53 42 16 77 3 41 84 96 55 67 54 40 57 31 23 62 74 27 21 59 52 25 38 44 80 18 6 49 92 47 48 46 30 74 2 27 47 33 24 46 30 1 85 73 16 99 82 73 74 27 89 59 23 73 16 50 53 1 85 11 64 81 84 96 55 67 54 57 93 86 53 1 66 89 29 28 7 87 9 43 74 2 89 98 53 1 85 53 42 16 77 3 41 84 96 55 67 54 40 57 93 31 59 23 99 82 73 74 27 47 59 33 90 35 94 32 64 65 6 49 41 84 96 55 67 54 40 57 93 70 14 76 0 30 99 82 73 70 52 25 94 32 64 85 88 18 27 47 33 24 46 30 1 85 11 64 60 70 14 \\
    \midrule
    \textbf{Speech (En)} & \makecell[c]{\includegraphics[width=0.99\linewidth]{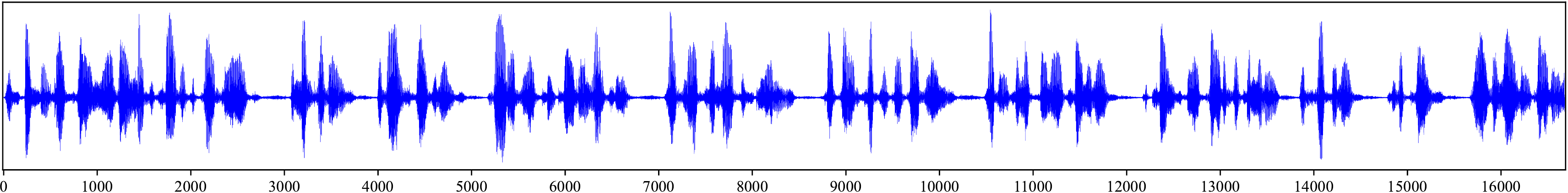}} \\
    \midrule
    \multirow{3}{*}{\textbf{ASR Output}} & The special airplane where they had to travel to Pagoric. Among others, the vice-presidents Dalibor, Kukir, and Rachel has not come out this morning for broad technical problems, having to look for an alternative to transport part of the committee executive to the place where the rating party was going to be held. \\
    \bottomrule
    \end{tabular}}
    \caption{Examples of synthetic En$\rightarrow$Es ST data.}
    \label{tab:case-enes}
\end{table*}

\begin{table*}[t]
    \centering
    \resizebox{\textwidth}{!}{
    \begin{tabular}{l|p{23cm}}
    \toprule
    \multicolumn{2}{c}{\textbf{\textit{Diverse} \textsc{BT4ST}: One German text$\rightarrow$Multiple English speeches}} \\
    \midrule
    \multirow{4}{*}{\textbf{Target (De)}} & Wegen der aufwendigen Ausstattung des Tunnels gehen die Ermittler davon aus, dass er von Architekten und Ingenieuren konstruiert wurde und dass der Bau rund ein Jahr in Anspruch nahm. \\ 
    & (\emph{Due to the elaborate configuration of the tunnel, investigators are working on the assumption that it was build by architects and engineers and that the construction took around one year.}) \\
    \midrule
    \multicolumn{2}{c}{\textbf{Output 1}} \\
    \midrule
    \multirow{6}{*}{\textbf{Unit}} & 71 47 11 45 64 74 27 89 78 33 24 61 68 9 29 28 23 73 16 99 82 11 64 53 73 74 27 78 14 76 53 16 74 2 50 42 44 80 85 73 16 99 82 73 62 74 27 31 59 33 68 44 80 85 19 81 83 84 57 96 55 39 67 54 57 93 82 11 45 64 53 44 80 18 66 77 87 9 43 6 49 92 31 23 73 1 66 89 94 32 64 85 5 30 29 28 23 73 16 50 87 91 9 1 21 95 23 42 44 80 18 82 87 9 85 53 60 70 14 76 62 29 28 92 57 89 23 44 80 18 6 49 92 21 52 24 61 43 2 31 23 53 1 66 47 24 13 58 37 24 46 30 74 2 89 23 62 74 27 31 59 87 91 9 43 74 2 6 49 7 23 44 80 85 87 38 44 80 18 21 95 23 44 80 26 11 45 64 0 79 29 28 49 41 84 57 96 55 39 67 54 57 86 38 44 80 82 87 9 74 2 82 73 74 27 89 23 44 80 18 6 49 92 52 24 68 43 74 2 95 23 44 80 85 73 16 77 73 52 70 52 25 91 17 44 18 85 11 98 45 64 0 79 81 83 20 \\
    \midrule
    \textbf{Speech (En)} & \makecell[c]{\includegraphics[width=0.99\linewidth]{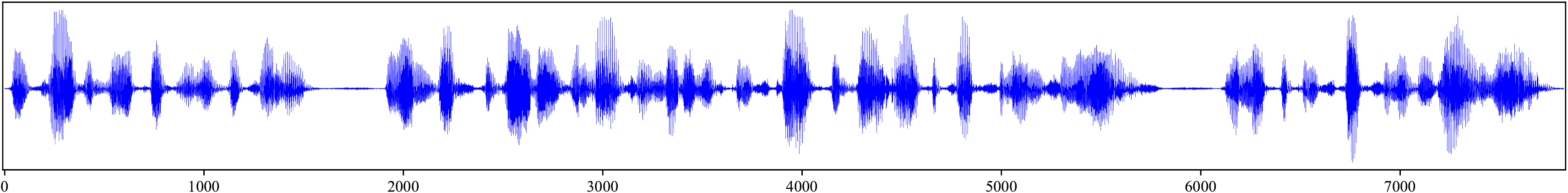}} \\
    \midrule
    \multirow{2}{*}{\textbf{ASR Output}} & Because of the equipment of the tunnel. The investigators imagined that it was constructed by architects and engineers, and that the construction of around a year. \\
    \midrule
    \multicolumn{2}{c}{\textbf{Output 2}} \\
    \midrule
    \multirow{6}{*}{\textbf{Unit}} & 71 47 45 74 27 89 59 33 68 9 29 28 23 73 16 99 66 82 73 74 27 47 59 33 91 43 6 15 7 23 62 1 21 95 92 27 31 59 33 68 44 80 85 19 90 35 73 74 27 78 14 76 53 9 74 2 50 42 44 80 18 2 31 41 84 57 96 55 39 67 54 57 93 82 11 45 64 53 44 80 18 66 77 87 91 43 6 15 49 92 31 23 1 66 89 87 94 32 64 85 5 30 29 28 60 70 52 25 94 32 64 29 28 92 82 87 9 85 53 60 70 14 76 62 29 28 92 57 89 23 44 80 18 6 49 92 21 52 24 61 43 74 2 31 23 53 1 66 47 24 13 58 37 24 46 30 74 2 89 23 62 74 27 31 59 87 91 43 74 2 6 15 7 23 44 80 26 87 38 44 80 18 21 95 23 44 80 26 11 0 79 29 28 41 84 57 96 55 39 67 54 57 93 86 38 44 80 82 87 9 74 2 82 73 74 27 89 23 44 80 18 6 49 92 21 52 24 68 43 74 2 65 95 23 44 80 60 70 14 76 62 29 28 60 70 14 68 44 80 18 98 0 79 81 83 20 \\
    \midrule
    \textbf{Speech (En)} & \makecell[c]{\includegraphics[width=0.99\linewidth]{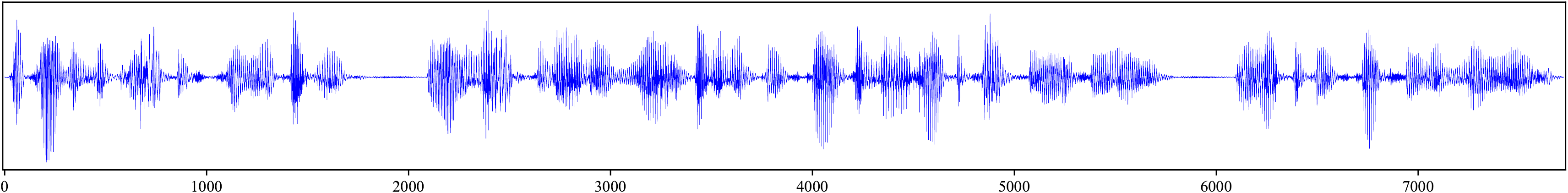}} \\
    \midrule
    \multirow{2}{*}{\textbf{ASR Output}} & Because of the passage tunnel equipment, the investigators raised that it was constructed by architects and engineers, and that the construction was one year. \\
    \midrule
    \multicolumn{2}{c}{\textbf{Output 3}} \\
    \midrule
    \multirow{5}{*}{\textbf{Unit}} & 71 47 11 45 64 74 27 89 59 33 68 9 29 28 23 73 16 99 82 11 64 53 73 74 27 78 14 76 53 9 74 2 50 42 44 18 2 31 23 73 16 99 82 73 62 74 27 31 59 33 68 44 80 85 19 81 83 84 57 96 55 39 67 54 57 93 82 11 45 64 53 44 80 18 66 77 87 9 43 6 49 92 31 23 73 74 27 89 59 94 32 64 85 5 30 65 6 49 7 69 16 50 18 29 28 92 82 87 9 85 53 60 70 14 76 62 29 28 92 89 23 44 80 18 6 49 92 52 24 68 43 74 2 31 23 53 1 66 47 24 13 58 37 24 46 30 74 2 89 23 62 74 27 31 59 87 91 43 74 2 6 49 7 23 44 80 85 87 38 44 80 18 21 95 23 44 80 26 11 0 79 29 28 49 41 84 57 96 55 39 67 54 57 86 38 44 80 18 82 87 9 74 2 82 73 70 52 25 91 17 68 80 18 85 23 73 16 66 77 87 38 44 18 21 95 60 90 35 11 64 74 27 31 59 94 32 74 27 89 23 53 44 80 85 73 11 98 45 0 79 81 83 20 \\
    \midrule
    \textbf{Speech (En)} & \makecell[c]{\includegraphics[width=0.99\linewidth]{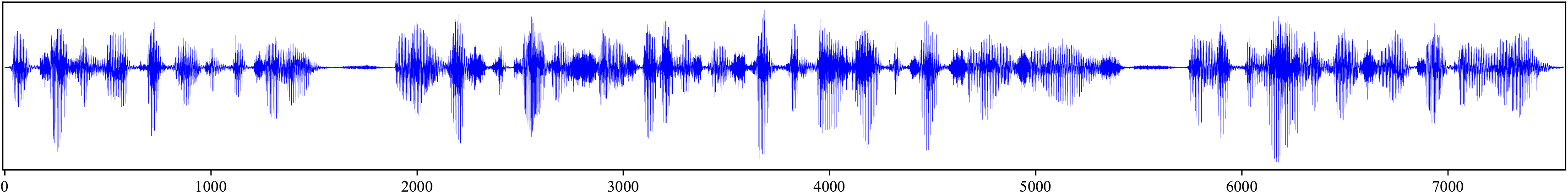}} \\
    \midrule
    \multirow{2}{*}{\textbf{ASR Output}} & Because of the equipment of the tunnel. The investigator assumes that it was constructed by architects and engineers and that the round eventually taken a year. \\
    \midrule
    \multicolumn{2}{c}{\textbf{Output 4}} \\
    \midrule
    \multirow{6}{*}{\textbf{Unit}} & 71 82 11 45 64 53 44 80 18 66 77 87 9 43 6 49 92 31 23 73 74 27 89 59 94 32 64 65 6 95 23 44 80 85 73 16 99 82 73 62 74 27 31 59 33 68 80 85 19 81 83 84 57 96 55 39 67 54 57 93 86 53 62 29 28 92 82 11 45 64 53 44 80 18 66 77 87 9 43 6 49 92 31 23 73 74 27 89 59 94 32 64 65 6 95 23 42 80 81 83 84 57 96 55 39 67 54 57 93 75 9 29 28 66 47 53 44 80 18 66 31 23 62 29 28 7 87 24 13 58 44 80 18 66 47 24 13 58 37 24 46 30 74 2 89 23 62 74 2 27 31 59 87 91 9 43 74 2 6 49 7 23 44 80 85 53 44 80 18 21 95 23 44 80 26 11 0 79 65 6 49 41 84 57 96 55 39 67 54 57 86 38 44 80 18 82 87 9 74 2 82 73 74 27 89 23 44 80 18 6 49 92 21 52 25 24 68 43 74 2 65 95 23 44 80 85 73 16 99 82 73 74 27 89 23 44 80 18 6 49 92 52 24 68 43 74 2 65 95 23 44 80 85 73 16 77 73 11 98 45 64 0 79 81 83 20 \\
    \midrule
    \textbf{Speech (En)} & \makecell[c]{\includegraphics[width=0.99\linewidth]{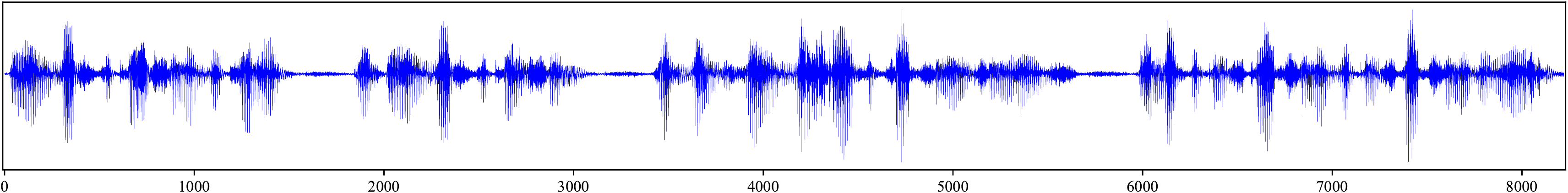}} \\
    \midrule
    \multirow{1}{*}{\textbf{ASR Output}} & The investigation of the tunnel is the investigation has been designed by architects and engineers and that the construction of the construction of a year. \\
    \bottomrule
    \end{tabular}}
    \caption{Examples of synthetic En$\rightarrow$De ST data generated by \textbf{\textit{Diverse} \textsc{BT4ST}} method.}
    \label{tab:case-diverse}
\end{table*}

\end{document}